\documentclass[journal]{IEEEtran}

\ifCLASSINFOpdf
\else
\fi

\usepackage{graphicx}
\usepackage{gensymb}
\usepackage{CJK}
\usepackage{booktabs}
\usepackage{multirow}
\usepackage[table,xcdraw]{xcolor}
\usepackage{hyperref}
\hypersetup{pagebackref=true,
linkcolor=black,
anchorcolor=black,
citecolor=black
}

\begin{document}
\title{Data Augmentation for Depression Detection  Using Skeleton-Based Gait Information}

\author{\IEEEauthorblockN{
		Jingjing Yang$^{1,\ast}$ ,
        Haifeng Lu$^{1,\ast}$ ,
        Chengming Li$^{2,\dagger}$,
        Xiping Hu$^{1,2,\dagger}$ ,
        Bin Hu$^{1,3,\dagger}$ 
	}                                     % ...
	\\
	\IEEEauthorblockA{$1.$% 1st affiliations
		School of information Science and Engineering, Lanzhou University, Lanzhou China.}
		\\
	\IEEEauthorblockA{$2.$% 2nd affiliations
		Shenzhen Institutes of Advanced Technology, Chinese Academy of Sciences, Shenzhen, China.}
		\\
	\IEEEauthorblockA{$3.$% 3rd affiliations
		Institute of Engineering Medicine, Beijing Institute of Technology, Beijing, China.}
	\\
	\IEEEauthorblockA{$\left\lbrace yangjj19, luhf18, huxp, bh \right\rbrace @lzu.edu.cn, cm.li@siat.ac.cn$}
	
	\thanks{$\dagger$ These authors are corresponding author.}
	\thanks{$\ast$ These authors have contributed equally to this work.}
}

\markboth{Journal of \LaTeX\ Class Files,~Vol.~XX, No.~XX, XXX~2021}%
{Shell \MakeLowercase{\textit{et al.}}: Bare Demo of IEEEtran.cls for IEEE Journals}



\maketitle

\begin{abstract}

In recent years, the incidence of depression is rising rapidly worldwide, but large-scale depression screening is still challenging. Gait analysis provides a non-contact, low-cost, and efficient early screening method for depression. However, the early screening of depression based on gait analysis lacks sufficient effective sample data. In this paper, we propose a skeleton data augmentation method for assessing the risk of depression. First, we propose five techniques to augment skeleton data and apply them to depression and emotion datasets. Then, we divide augmentation methods into two types (non-noise augmentation and noise augmentation) based on the mutual information and the classification accuracy. Finally, we explore which augmentation strategies can capture the characteristics of human skeleton data more effectively. Experimental results show that the augmented training data set that retains more of the raw skeleton data properties determines the performance of the detection model. Specifically, rotation augmentation and channel mask augmentation make the depression detection accuracy reach $92.15$\% and $91.34$\%, respectively.

\end{abstract}


% \fi
\IEEEpeerreviewmaketitle

\section{Introduction}
\label{intro}

According to the data from the World Health Organization (WHO), there are more than 340 million people worldwide suffering from depression, which has become the second cause of death after cancer \cite{WHO}. In 2020, the Chinese government proposed mental health services for high-risk crowds, such as teenagers, pregnant women, the elderly, especially depression screening was included in student's physical examination \cite{NHC}. However, due to the shortage of professional psychiatric doctors and social discrimination associated with mental disorders, large-scale screening for depression is still a challenge. Fortunately, Artificial Intelligence (AI) has made it possible for large-scale depression screening \cite{branco2018identification,cohn2018multimodal,ringeval2019avec}.

Some scholars pointed out that there is convincing evidence of bidirectional interactions between depression and gait information \cite{sanders2010gait}. Hence, some scholars pay attention to automatic depression detection using gait analysis integrated with AI technologies \cite{fang2019depression,zhao2019see,yuan2018depression}. However, there is a common problem that lack sufficient labeled datasets. This problem is affected by objective factors (time, cost), and is difficult to solve in a short time. According to relevant papers, small sample datasets are more likely to overfit during training \cite{shorten2019survey}. Therefore, we need to consider how to avoid overfitting in the limited data and improve the performance of the depression detection model as far as possible.

Data augmentation includes a set of techniques that can augment the size and quality of training datasets, and it has been proved to be effective to improve the performance of the deep learning models \cite{hussain2017differential}. Data augmentation methods can obtain a more complex representation of the raw data, reduce the gap between the training set and test set, and enable the neural network to better understand the distribution of data on the dataset. Existing data augmentation technologies are divided into two aspects: data warping and oversampling \cite{Connor2019A}. Data warping augmentations convert existing images, including  geometric and color transformations, random erasing, adversarial training, and neural style transfer. Oversampling augmentations create synthetic instances and add them to the training set, including mixing images, feature space augmentations, and the Generative Adversarial Networks (GAN) \cite{goodfellow2014generative}.

Differential data augmentation techniques have been proposed in computer vision, including Random Erasing \cite{zhong2017random}, AutoAugment \cite{cubuk2018autoaugment}, GridMask \cite{chen2020gridmask}. Although different augmentation strategies and their combinations have been investigated widely for image recognition tasks, the augmentation strategies for skeleton-based gait data are rarely studied. The lack of gait training data makes it critical to explore data augmentation strategies.

In this work, we preprocess the skeleton-based gait data collected by the Kinect camera firstly. Then, we propose five augmentation methods for skeleton-based gait data. Thirdly, we apply five augmentation methods to the postgraduate depression gait dataset and use two deep learning models (Long-term Short-Term Memory (LSTM) \cite{hochreiter1997long} and Temporal Convolutional Network (TCN) \cite{lea2016temporal}) to test our method. Finally, we verify the experimental results on the public emotion dataset and analyze the augmentation effects of these methods based on the mutual information between the augmented data and the raw data. The results show that the augmentation methods can be divided into two types: non-noise augmentation and noise augmentation. The effect of non-noise augmentation is obviously better than that of noise augmentation. Rotation and channel mask augmentations increase the classification accuracies to $92.15$\% and $91.34$\% respectively.

In summary, we make the following contributions:

$\bullet$ $\ $ In order to solve the problem of insufficient sample gait data, five data augmentation methods (rotation, shear, Gaussian noise, channel mask, and joint mask) are proposed. Experimental results show that the augmentation strategy has a significant impact on the depression detection models, the rotation and channel strategies increase the accuracy by about $6$\%.

$\bullet$ $\ $ By calculating the mutual information between the augmented data and the raw data, the proposed augmentation methods on skeleton-based gait data are divided into two types: non-noise augmentation and noise augmentation. These two types can preliminary estimate the results of data augmentation.

$\bullet$ $\ $ We analyze the impact of five data augmentation methods on classification accuracy and conclude that augmented training data set that retains properties more of the raw skeleton data properties determines the performance of the model. This provides an idea for the application of data augmentation for depression detection using gait analysis in the future.

The rest of this paper is organized as follows: Sec. \ref{section:related work} introduces the research status of depression recognition and the background of data augmentation. Two gait datasets and five proposed data augmentation methods are depicted in Sec. \ref{section:method}. Sec. \ref{section:experimental} presents the experimental results and gives a comprehensive discussion of the proposed methods combined with the mutual information. The conclusion of this paper described in Sec. \ref{conclusion}.

\section{Related Work} 
\label{section:related work}
\subsection{Depression recognition and gait}
At present, the clinical diagnosis of depression is mainly based on symptoms, mental and scale examinations by psychiatrists \cite{beck2009depression}. However, most depression patients do not ask for help in time because patients usually cannot realize that they have depression in the early stages of depression. In recent years, many scholars recognize depression from the perspective of facial expression, voice, physiological signal, video, \emph{etc.} \cite{zhou2018visually,li2018improvement,shen2020optimal}. However, the detection of physiological signals requires a accurate detection system in a restrictive environment; Facial expression and voice collection need emotional stimulation in a specific environment. Therefore,large-scale depression detecting is still a difficult problem.

% In recent years, many scholars recognize depression from the perspective of facial expression, voice, physiological signal, video, \emph{etc.} In \cite{zhou2018visually}, the authors constructed a deep regression network based on the Depression Activation Map (DAM) generated by facial depression data, and recognized the salient areas of the input image according to its severity score. Li \emph{et al.} encoede depression-related features in the vocal tract and provided a more comprehensive audio representation to recognize depression \cite{li2018improvement}. In \cite{shen2020optimal}, the authors proposed an optimal channel selection method for EEG-based depression detection via Kernel-Target Alignment (KTA). But the detection of physiological signals requires a accurate detection system in a noise-free environment, facial expression and voice collection need emotional stimulation in a specific environment.

Research shows that emotion is interlocked with perception, cognition, motivation, and action \cite{Pessoa2017A}. Many scholars have studied the relationship between gait and emotion, e.g., happy walkers display an increase in gait speed, step length, arm swing, \emph{etc.}; Sad walkers display a decrease in gait speed, step length, arm swing, \emph{etc.} \cite{Gross2012Effort,Kang2016The,Michalak2009Embodiment,10.1007/978-3-540-74889-2_9}. Long-term exposure to negative emotions may lead to changes in brain areas related to perception, cognition, and motor systems \cite{2015Qualia}.

In some studies on the gait of people with depression, the researchers found that depression patients have a reduced gait velocity, vertical head movement, arm swing, stride length, \emph{etc.} \cite{lemke2000spatiotemporal,Sanders2016Gait}. In \cite{wang2020gait}, Wang \emph{et al.} extracted a novel Time-domain and Frequency-domain feature (TF-feature) and a Spatial Geometric feature (SG-feature), and investigated the effectiveness of fused features of gait data for the non-contact depression detection. A rigid-body representation of the human body was proposed in \cite{Luhealth}, which can improve the robustness of detecting depression patients with noisy input and reduce the classification time. %In \cite{LuIGBH}, a new gait feature called Joint Energy feature(JE-feature) was proposed, which can be used as an important indicator to assess the risk of depression. 

In previous studies, scholars have been committed to exploring new gait features and using traditional machine learning models for classification, and the accuracy has been continuously improved. However, traditional machine learning technologies were limited in their ability to process natural data in their raw form, and lack the ability to express complex functions, making it difficult to solve more complex natural signal processing problems. Deep learning allows computational models that are composed of multiple processing layers to learn representations of data with multiple levels of abstraction, and demonstrate a powerful ability to learn the essential characteristics of a data set from a small number of samples \cite{lecun2015deep}. At present, few people applies deep learning methods to gait depression recognition.

\subsection{Data augmentation}

In \cite{wong2016understanding}, the authors pointed out data augmentation can be divided into augmenting raw data and feature space. After experimental comparison, it is found that augment raw data is better. Therefore, we focus on the raw data which is three-dimensional skeleton data collected by Kinect. In this paper, geometric transformation and random erasure are mainly used. Geometric transformation includes three methods: rotation, shearing, and adding Gaussian noise. Random erasure refers to hiding some random or specific point data. Of course, the use of GAN for style transfer and feature space data augmentation is also a problem worthy of further study.

Rotation, noise injection and random erasing are commonly image augmentation methods. Rotation augmentation is accomplished by rotating the image right or left on an axis between 1$\degree$ and 359$\degree$ \cite{shorten2019survey}. Noise injection consists of injecting a matrix of random values usually drawn from a Gaussian distribution \cite{shorten2019survey}. Moreno-Barea \emph{et al.} confirmed that adding noise to the images can help CNN learn more robust features by testing on 9 datasets \cite{moreno2018forward}. Random erasing \cite{zhong2020random} is a data augmentation technique developed by Zhong \emph{et al.}, which can be understood as losing part of the data information before entering the network. Random erasing forces the model to learn more descriptive features about the rest of the image, thereby preventing the model from overfitting certain visual features of the image. In this paper, we use two random erasing strategies: joint mask and channel mask.

%\subsubsection{Neural Networks-based data augmentation}

%\ 

Variational auto-encoders (VAE) \cite{kingma2013auto} have become a popular method of unsupervised learning. They are constructed by neural networks and can be trained with stochastic gradient descent \cite{doersch2016tutorial}. VAE has shown good prospects for generating complex data including handwritten digits, faces, physical models of house numbers, segmentation, and predicting the future from static images.

GAN architecture proposed by Ian Goodfellow firstly \cite{goodfellow2014generative} is a framework for generative model through adversarial training. GAN architecture is composed of a generative model and a discriminative model that antagonize each other. In the most ideal state, the generator can generate images that are difficult for the discriminator network to judge "real and false". The excellent performance of GAN has made people pay more and more attention to how to apply it to data augmentation. \cite{karras2019style,bau2018gan,zhu2017unpaired}.

\section{Method} \label{section:method}

\subsection{Dateset}  
\begin{figure}[htbp]
  \centering
  {\includegraphics[width=3.5in]{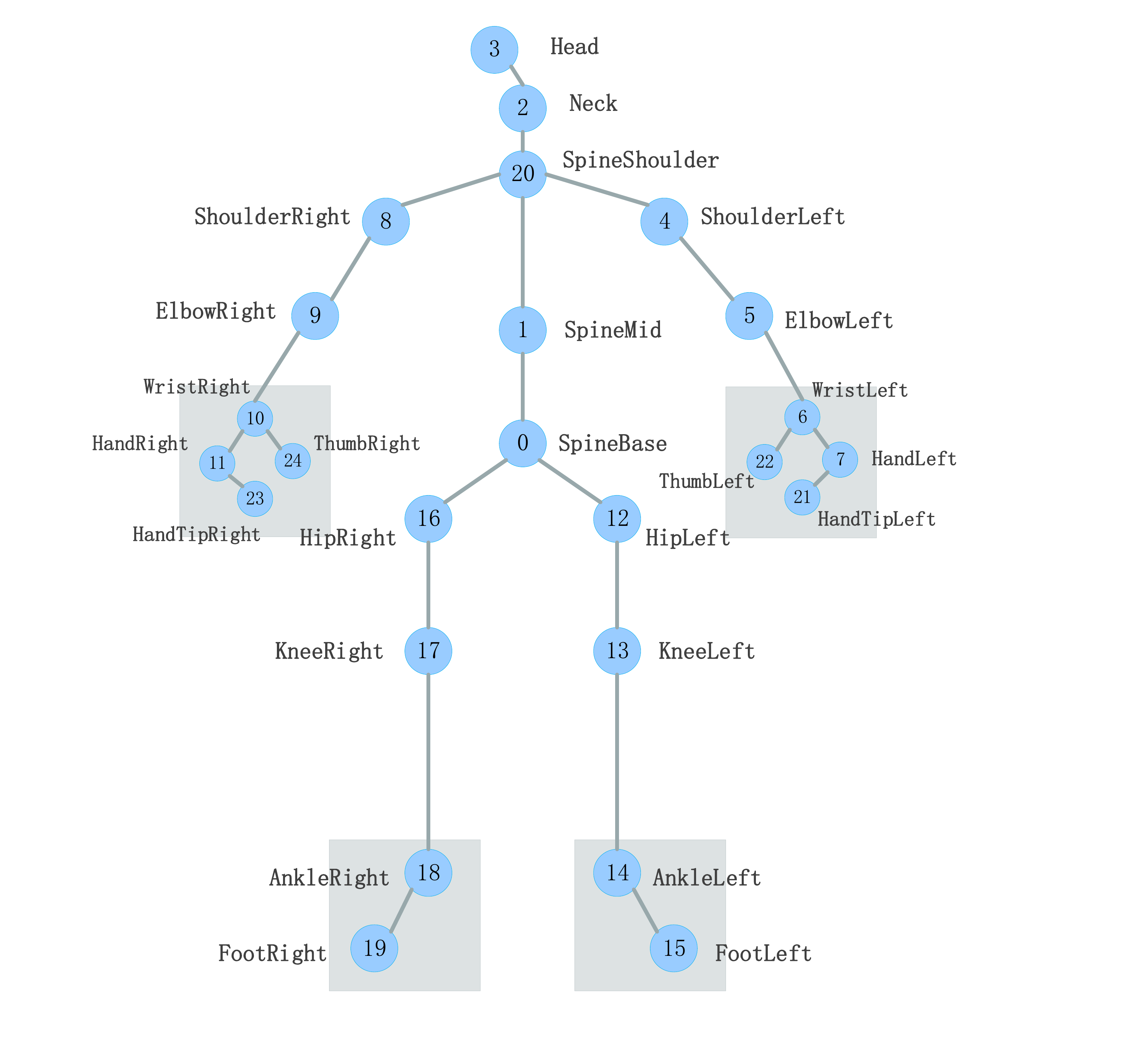}}
  \caption{The diagram of joints location \cite{Kinect}.}
  \label{joint}
 \end{figure}

\subsubsection{Depression dataset}

\ 
In this paper, we use the skeleton coordinate data of 95 postgraduate students aged $22$ to $18$ \cite{Luhealth}, including 43 scored-depressed students and 52 non-depressed students based on the scores of  the Patient Health
Questionnaire (PHQ-9, Chinese version) \cite{kroenke2001phq} and Zung Self-
rating Depression Scale (SDS, Chinese version) \cite{zung1965self}. All participants were asked to walk forth and back twice on a 10-meter-long path at a comfortable speed and posture. Two Microsoft Kinect V2 cameras were used to obtain data of 25 joint points (as shown in Fig. \ref{joint}) of the human body. 

\subsubsection{Emotion dataset}

\ 

We use the part of the emotional multimodal database that records the skeletal points of the human body \cite{sapinski2018multimodal}. During the collection process, all actors were required to perform emotional states separately in the following order: neutral, sadness, surprise, fear, disgust, anger, and happiness. Without any guidance or prompts from the researcher, everyone performed each emotion five times. The Kinect V2 sensor was used to capture the skeleton data of 25 joint points of the human body. Except for some poor quality data, the emotional skeleton database used in this study finally contains 474 samples of 13 participants. 

\subsection{Data preprocessing method}
\subsubsection{Depression dataset}

\

The preprocess of the depression dataset includes the following four processes: data filtering, angle transformation, de-noise and re-sampling, and data simplification.

\noindent Data Filtering: During the collection, the device estimated the coordinates of the skeleton joints from the front view and the back view. In the first step, we segment them into two streams of data. According to previous reports, the estimated skeleton in the back view is not as accurate as the skeleton in the front view. Therefore, we only analyze the skeleton in the front view. The second step is to observe whether the target skeleton presents a good shape and delete the data of limbs deformation.

\noindent Angle Transformation: Because the camera has an angle to the ground, participants do not walk horizontally along the Z axis in the Kinect coordinate system. System errors are introduced when filtering Z-axis data. In order to analyze the data accurately, the coordinate system of the raw data must be converted. We perform coordinate transformation on all recorded data according to the transformation matrix in Eq. \ref{preprocessing_1}.

\begin{equation}
\label{preprocessing_1}
\left[
\begin{array}{cccc}
x^{\prime}\ y^{\prime}\ z^{\prime}
\end{array}
\right ]
=
\left[
\begin{array}{cccc}
x\ y\ z
\end{array}
\right ]
\left[
\begin{array}{cccc}
1& 0& 0\\ 
0& \cos\theta& \sin\theta\\ 
0& -\sin\theta& \cos\theta\\ 
\end{array}
\right ]
\end{equation}
where $[x^{\prime}\ y^{\prime}\ z^{\prime}]$ is the joint position after coordinate transformation; $[x\ y\ z]$ is the raw data; $\theta$ is the angle between the Kinect camera and the ground, calculated by the formula in \cite{Luhealth}.

In the Kinect coordinate system, each participant's walking trajectory is different. In order to eliminate the influence of absolute position information, the skeleton coordinates must be projected from the sensor space to the local space of the human body, and the center of the local space is located on the SpineBase joint (joint $0$ in Fig. \ref{joint}). The formula is shown in Eq. \ref{preprocessing_2}. As a result, vectors containing the position and direction of each joint relative to the Spinebase joint are obtained.

\begin{equation}
\label{preprocessing_2}
norm\_joint = joint - spinebase
\end{equation}

\noindent De-noise and Re-sampling: The Kinect camera is affected by different factors such as light and colors of the subject's clothes, which generate noise in the collected data. We use Gaussian filter \cite{gwosdek2011theoretical} to smooth data, reduce high-frequency noise, and increase identifiability \cite{fang2019depression,zhao2019see}. The length of the Gaussian filter window is $5$, and the convolution kernel is 1/16 $\times$ [1,4,6,4,1]. It filters the x-axis, y-axis and z-axis data separately based on the timestamp.

\noindent Data Simplification: During the process of data collection, participant‘s hand joints are prone to misidentification and there is almost no  movement among the fingers when walking \cite{sapinski2018multimodal}. In order to reduce the impact of error data on the training model, the average value of the left hand (joints $6,7,21,22$ in Fig. \ref{joint}) or the right hand (joints $10,11,23,24$ in Fig. \ref{joint}) is taken as the human hand coordinate. In addition, since the Kinect hanging on the ceiling will cause distortion of the recognition of the joints of the foot, we take the average value instead of them respectively. After data simplifying, the number of skeleton joints is changed from $25$ to $17$. This part is only suitable for rotation, adding Gaussian noise and shear augmentations.

\subsubsection{Emotion dataset}

\

Firstly, similar to the coordinate transformation method mentioned above, raw Kinect V$2$ data is affected by the distance between the actor/actress and the sensor during recording. Skeleton coordinates must be projected from the sensor space to the local space of the human body. Secondly, in order to unify the joint position values between higher and lower individuals, we use normalisation based on the distance between two joints  with the lowest position noise in all records: SpineBase (joint $0$ in Fig. \ref{joint}) and SpineShoulder (joint $20$ in Fig. \ref{joint}). The normalization process follows Eq. \ref{preprocessing_3}.

\begin{equation}
\label{preprocessing_3}
d_i = \frac{J_i}{J_{20}-J_0},
\end{equation}
where $d_i$ is the distance vector between the i and $J_0$ joints. ${J_{20}-J_0}$ refers to the difference between the individual joints $J_{20}$ (SpineShoulder) and $J_0$ (SpineBase) in the neutral state. $J_i$ refers to the joint vector of each joint point in the seven emotional states.

\subsection{Rotation and translation}
\begin{figure}[htbp]
  \centering
  \centerline{\includegraphics[width=2.3in]{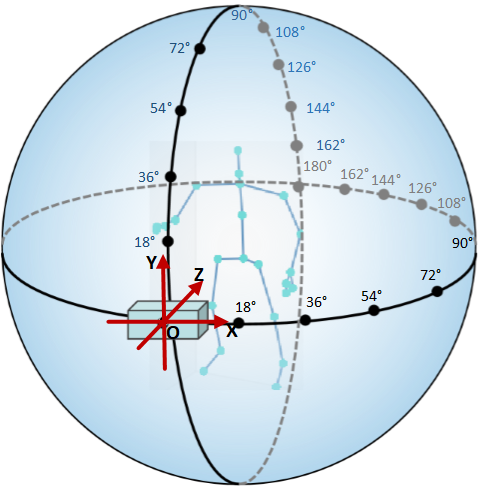}}
  \caption{The schematic of rotation augmentation.}
  \label{ro_sp}
 \end{figure}

Through the conversion of the original data, we obtain the skeleton data under different virtual cameras. The schematic is shown in Fig. \ref{ro_sp}, where the black dots are the locations of the virtual camera in different views. The specific operation steps are as follows. We take the participant as the center, and suppose the camera rotates and translates around the participant. The  rotation angle refers to the position of cameras in Chinese Academy of Sciences Institute of Automation (CASIA) dataset \cite{yu2006framework}, increasing by 18$\degree$ each time.
 
\begin{figure}[htbp]
  \centering
  \centerline{\includegraphics[width=3.5in]{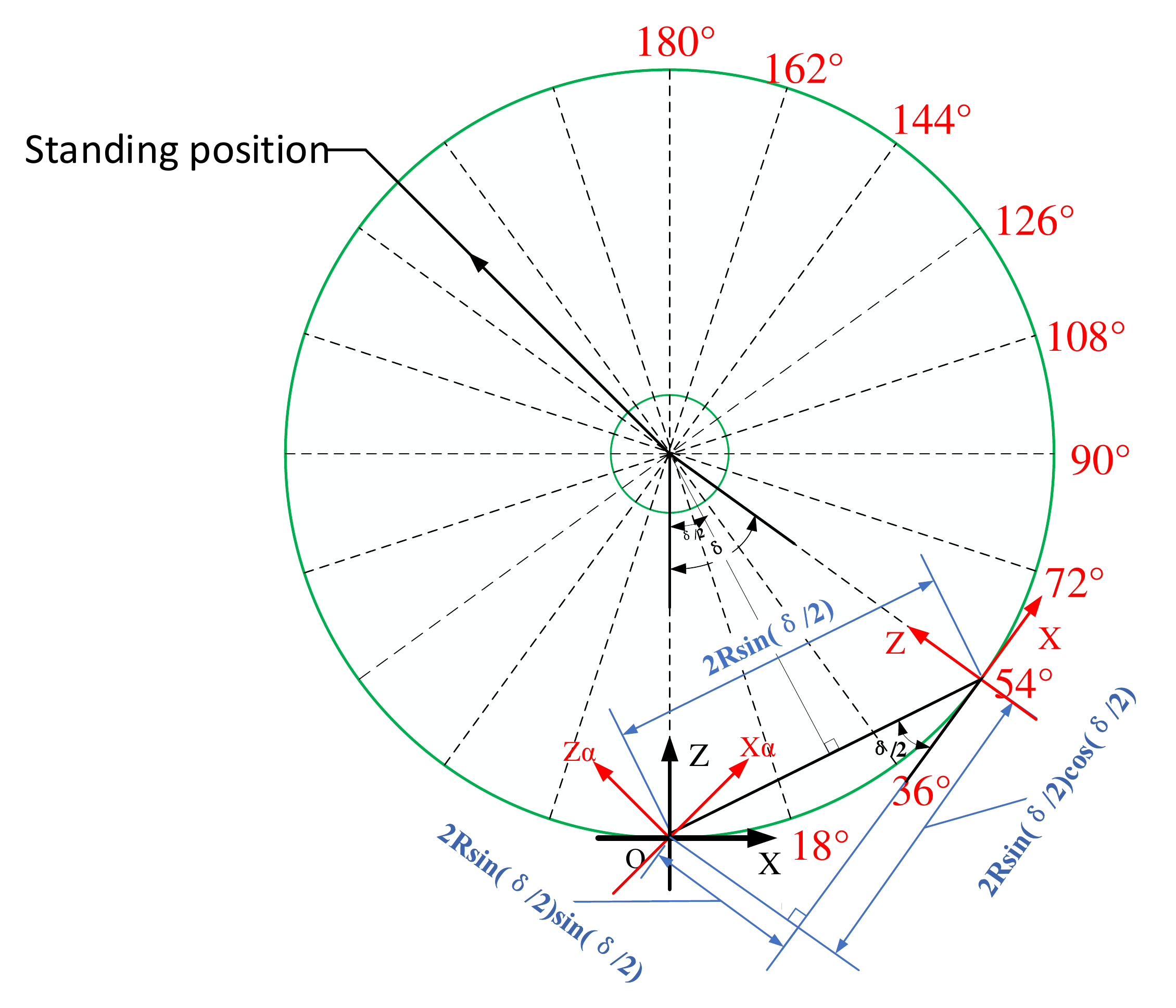}}
  \caption{The planform of camera position rotation.}
  
  \label{rotation-eps}
\end{figure}

We rotate the camera to the same angle with CASIA dataset. According to Eq. \ref{roation_max_right_left}, the camera is rotated horizontally to the position of '$X_{\alpha}OZ_{\alpha}$' in Fig. \ref{rotation-eps}. Since the camera revolves horizontally with the participant's SpineBase joint as the center, the change of y-axis is not involved. When rotating in the vertical direction, the \emph{ZOY} plane also performs the same operation without changing the x-axis. The formula is shown in Eq. \ref{roation_max_up_down}.

\begin{equation}
\label{roation_max_right_left}
\left[ 
\begin{matrix}
{x}_{\alpha} \ {y}_{\alpha} \ {z}_{\alpha} 
\end{matrix}
\right] 
=
\left[
\begin{matrix}
{x}^{\prime} \  {y}^{\prime} \  {z}^{\prime}
\end{matrix}\right]
\left[\begin{array}{ccc}   \cos \delta & 0 & -\sin \delta \\ 0 & 1 & 0\\ \sin \delta & 0 & \cos \delta\end{array}\right]
\end{equation}

\begin{equation}
\label{roation_max_up_down}
\left[ 
\begin{matrix}
{x}_{\beta}\ {y}_{\beta}\  {z}_{\beta}
\end{matrix}
\right] 
=
\left[
\begin{matrix}
{x}^{\prime}\ {y}^{\prime}\ {z}^{\prime}
\end{matrix}\right]
\left[\begin{array}{ccc}   1 & 0 & 0 \\ 0 & \cos \delta & \sin \delta\\ 0& -\sin \delta & \cos \delta\end{array}\right]
\end{equation}
where $\delta$ is the rotation angle, [$\begin{matrix}{x}^{\prime}\ {y}^{\prime }\ {z}^{\prime }\end{matrix}$] is the origin coordinate; [$\begin{matrix}{x}_{\alpha }\ {y}_{\alpha}\ {z}_{\alpha} \end{matrix}$] and 
[$\begin{matrix}{x}_{\beta }\ {y}_{\beta}\ {z}_{\beta} \end{matrix}$] are the coordinates after horizontal and vertical rotation respectively.

According to the scenario in Fig. \ref{ro_sp}, we need to translate the data. First, set the distance between the camera and the human body to $3$. Second, the displacement distance of the origin is obtained according to the geometric relationship in \ref {rotation-eps}. The calculation process is shown in Eq. \ref{tran_xz} and Eq. \ref{tran_yz}. For example, when the angle is equal to 54$\degree$, the coordinate system $X_{54}OZ_{54}$ in Fig. \ref{rotation-eps} is the position \emph{XOZ} after the rotation and translation. Finally, the camera is rotated and translated in the \emph{XOZ} plane and the \emph{YOZ} plane.

\begin{equation} \label{tran_xz}
\left\lbrace \begin{array}{l}
{x}^{\prime \prime}={x}_{\alpha}-2R\sin\left(\frac{\delta}{2} \right) \cos\left(\frac{\delta}{2} \right) \\ 
{y}^{\prime \prime}={y}_{\alpha} \\ 
{z}^{\prime \prime}={z}_{\alpha}+2R\sin\left(\frac{\delta}{2} \right) \sin\left(\frac{\delta}{2} \right)
\end{array}
\right. 
\end{equation}

\begin{equation}
\label{tran_yz}
\left\lbrace \begin{array}{l}
{x}^{\prime \prime}={x}_{\beta } \\ 
{y}^{\prime \prime}={y}_{\beta }-2R\sin\left(\frac{\delta}{2} \right) \cos\left(\frac{\delta}{2} \right) \\ 
{z}^{\prime \prime}={z}_{\beta }+2R\sin\left(\frac{\delta}{2} \right) \sin\left(\frac{\delta}{2} \right)
\end{array}
\right. 
\end{equation}

\begin{figure}[htbp]
	\centering
	%\begin{tabular}{cc}
	\begin{minipage}{0.3\linewidth}
		
		\centerline{\includegraphics[width=2.5cm,height=3.5cm, trim=50 35 220 60,clip]{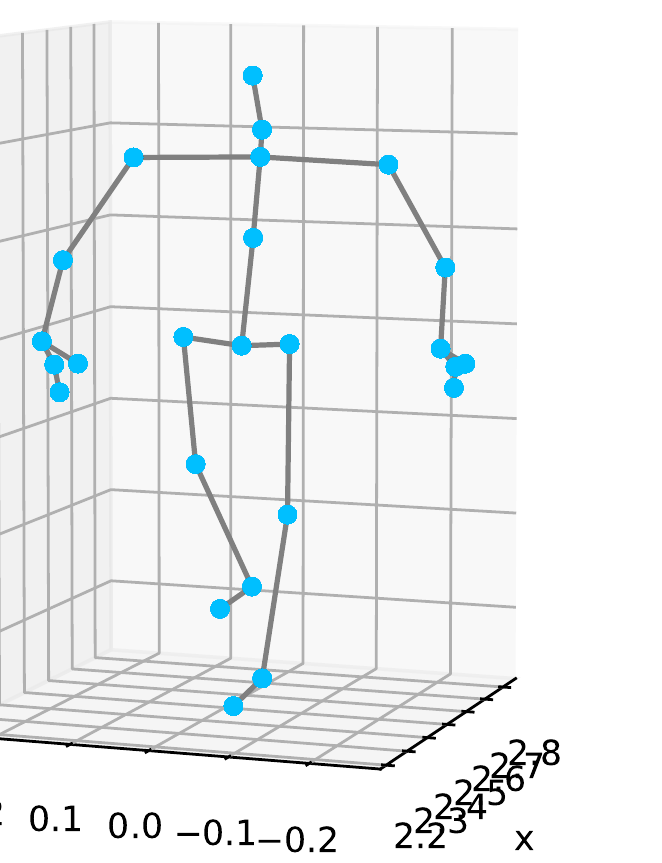}}
		\centerline{(a) raw}
	\end{minipage}
	\hfill
	\begin{minipage}{.3\linewidth}
		\centerline{\includegraphics[width=2.5cm,height=3.5cm, trim=50 35 220 60,clip]{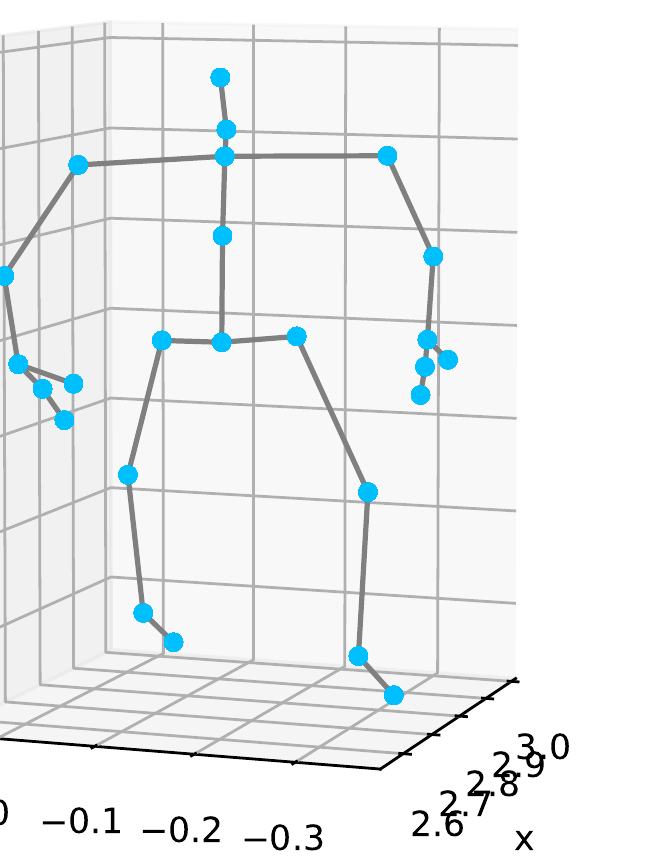}}
		\centerline{(b) raw-36-h}
	\end{minipage}
	\hfill
	\begin{minipage}{.3\linewidth}
		\centerline{\includegraphics[width=2.5cm,height=3.5cm, trim=50 35 220 60,clip]{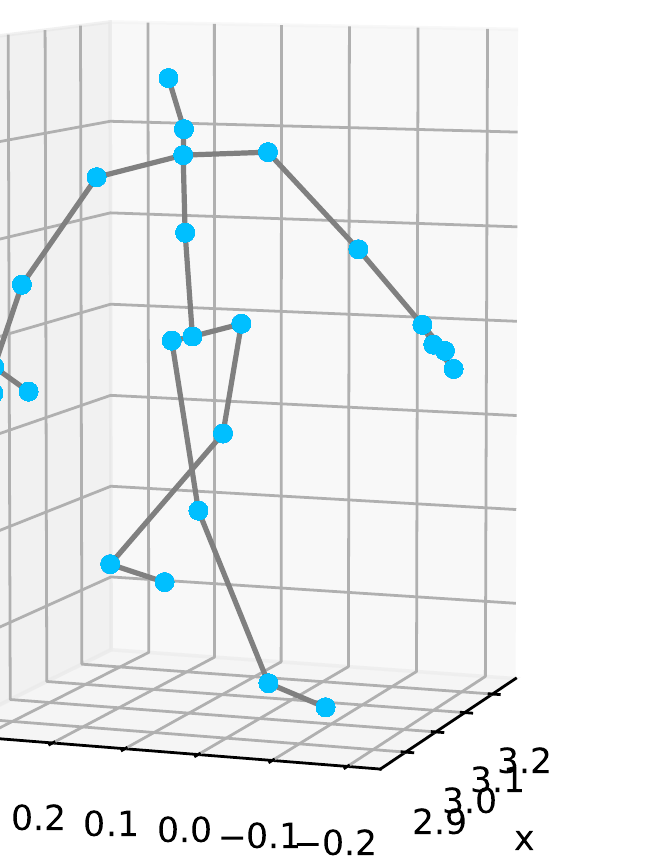}}
		\centerline{(c) raw-72-h}
	\end{minipage}
	\vfill
	\begin{minipage}{.3\linewidth}
		\centerline{\includegraphics[width=2.5cm,height=3.5cm, trim=50 35 220 60,clip]{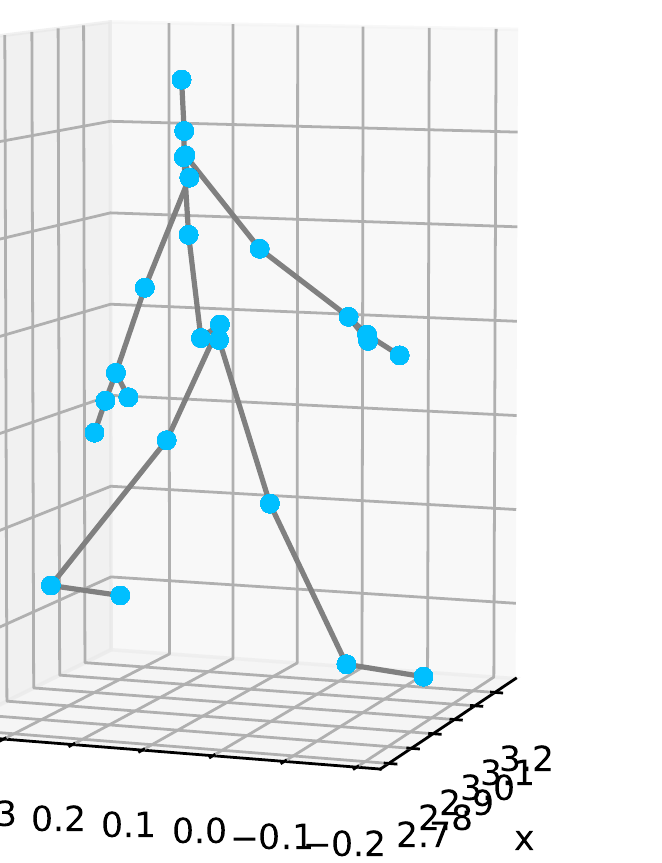}}
		\centerline{(d) raw-108-h}
	\end{minipage}
	\hfill
	\begin{minipage}{0.3\linewidth}
		\centerline{\includegraphics[width=2.5cm,height=3.5cm, trim=50 35 220 60,clip]{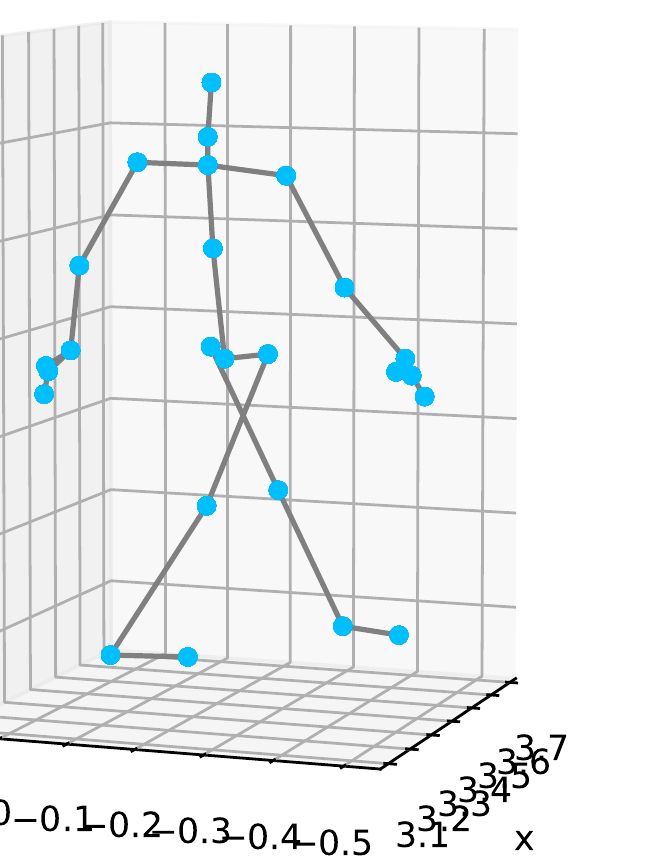}}
		\centerline{(e) raw-144-h}
	\end{minipage}
	\hfill
	\begin{minipage}{0.3\linewidth}
		\centerline{\includegraphics[width=2.5cm,height=3.5cm, trim=50 35 220 60,clip]{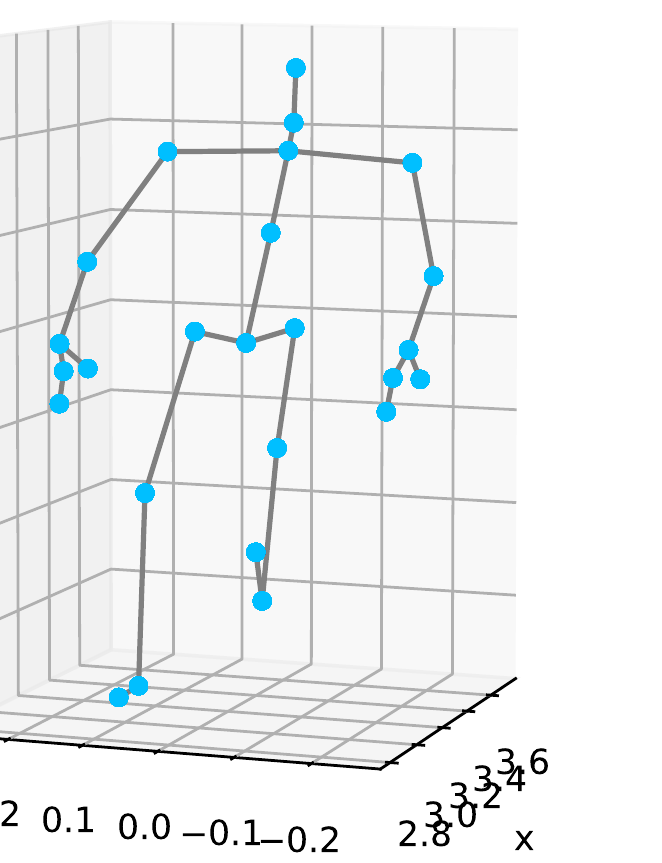}}
		\centerline{(f) raw-36-v}
	\end{minipage}
	\vfill
	\begin{minipage}{0.3\linewidth}
		\centerline{\includegraphics[width=2.5cm,height=3.5cm, trim=50 35 220 60,clip]{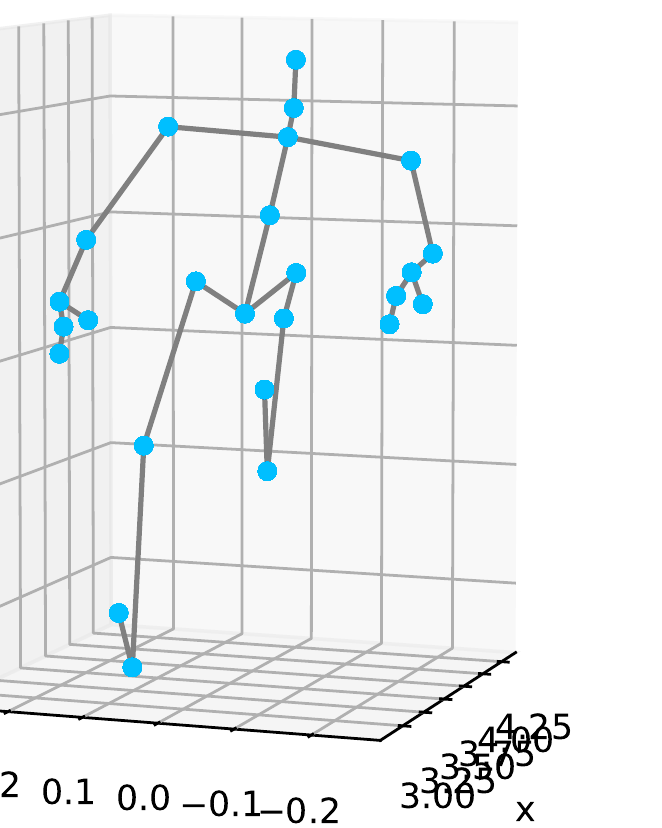}}
		\centerline{(g) raw-72-v}
	\end{minipage}
	\hfill
	\begin{minipage}{0.3\linewidth}
		\centerline{\includegraphics[width=2.5cm,height=3.5cm, trim=50 35 220 60,clip]{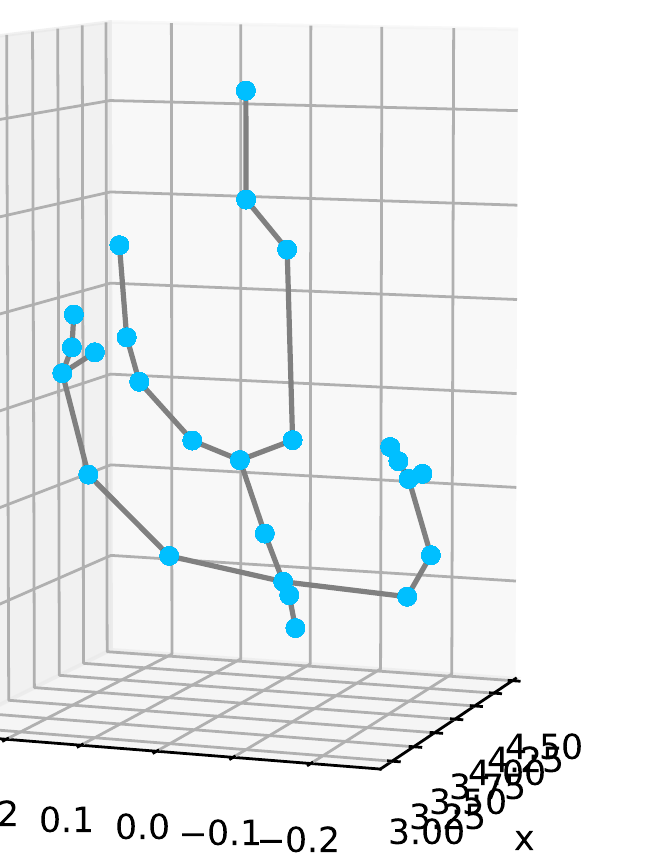}}
		\centerline{(h) raw-108-v}
	\end{minipage}
\hfill
	\begin{minipage}{0.3\linewidth}
		\centerline{\includegraphics[width=2.5cm,height=3.5cm, trim=50 35 220 60,clip]{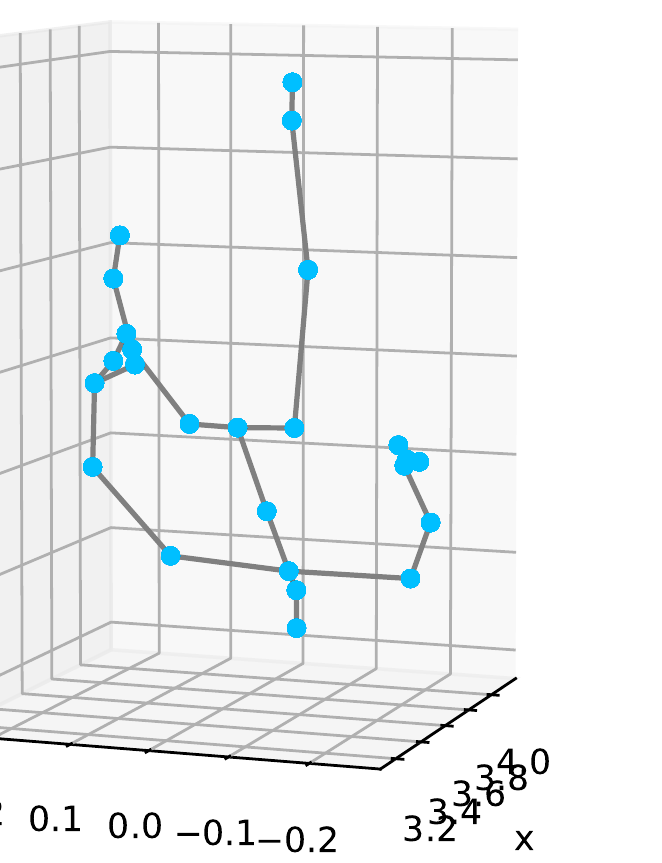}}
		\centerline{(h) raw-144-v}
	\end{minipage}

	%\end{tabular}
	\caption{Visualization of rotation augmentation.}
	\label{fig:res}
\end{figure}

After rotation and translation, we choose different $\delta$ to augment the data. The angles of 18$\degree$, 36$\degree$, 54$\degree$, 72$\degree$, 90$\degree$, 108$\degree$, 126$\degree$, 144$\degree$, 162$\degree$, and 180$\degree$ are transformed in horizontal and vertical directions. The visualization pictures are shown in Fig. \ref{fig:res}. Fig. \ref{fig:res}. a) is the raw data, and Fig. \ref{fig:res}. b) - Fig. \ref{fig:res}. e) correspond to the transformation results of 36$\degree$, 72$\degree$, 108$\degree$, and 144$\degree$ respectively in the horizontal direction. Fig. \ref{fig:res}. f) - Fig. \ref{fig:res}. h) correspond to the transformation results of 36$\degree$, 54$\degree$, 108$\degree$, and 144$\degree$ respectively in the vertical direction.
\subsection{Shear}
\begin{figure}[htbp]
	\centering
	%\begin{tabular}{cc}
	\begin{minipage}{0.3\linewidth}
		\centerline{\includegraphics[width=2.5cm,height=3.5cm, trim=20 35 220 60,clip]{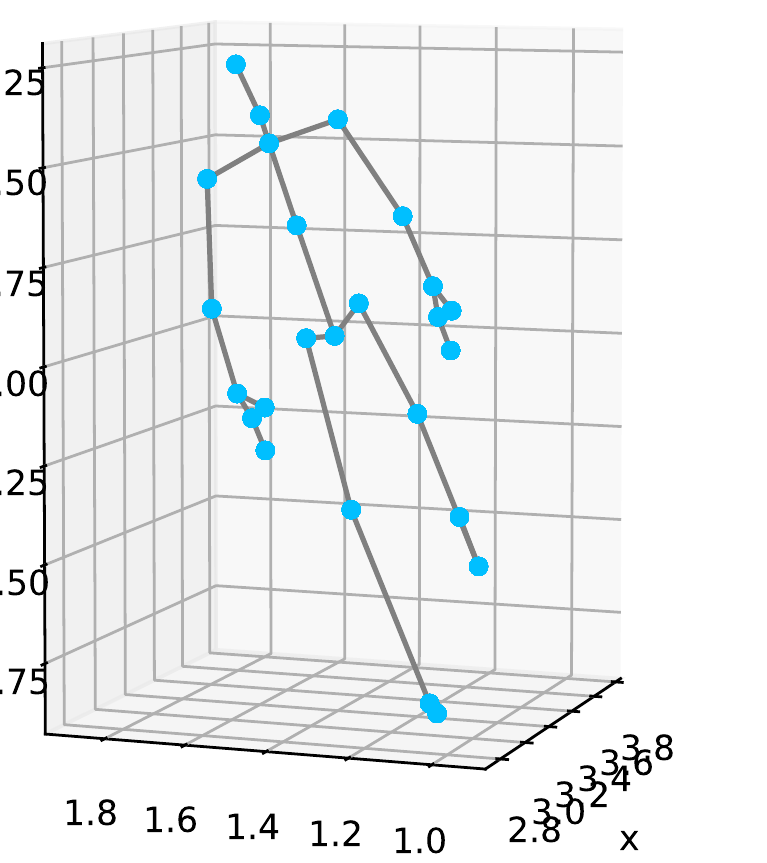}}
		\centerline{(a) shear-1}
	\end{minipage}
	\hfill
	\begin{minipage}{.3\linewidth}
		\centerline{\includegraphics[width=2.5cm,height=3.5cm, trim=20 35 220 60,clip]{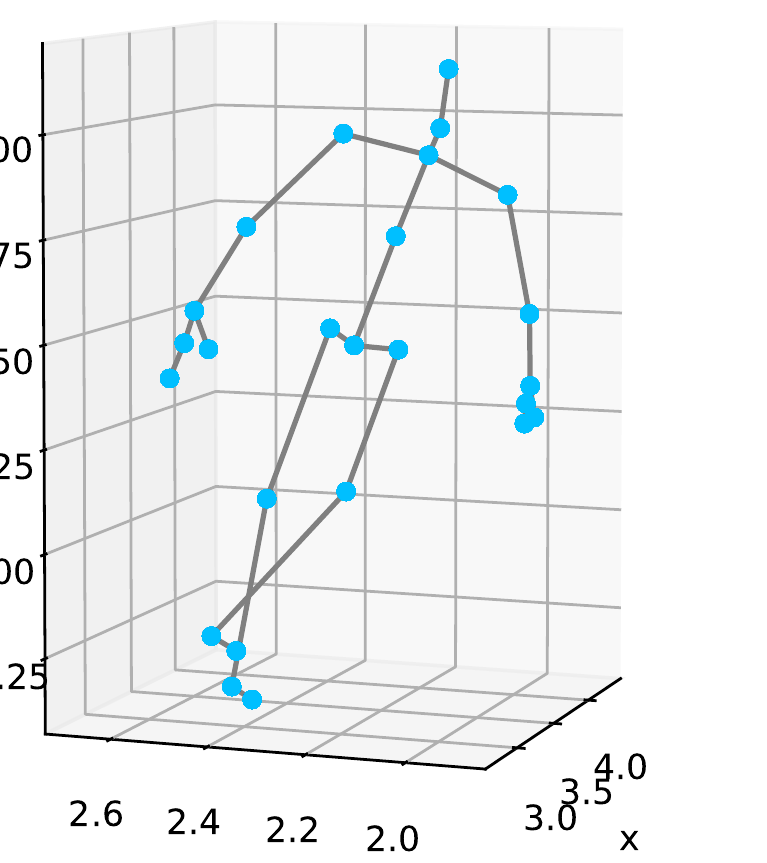}}
		\centerline{(b) shear-2}
	\end{minipage}
	\hfill
	\begin{minipage}{.3\linewidth}
		\centerline{\includegraphics[width=2.5cm,height=3.5cm, trim=20 35 220 60,clip]{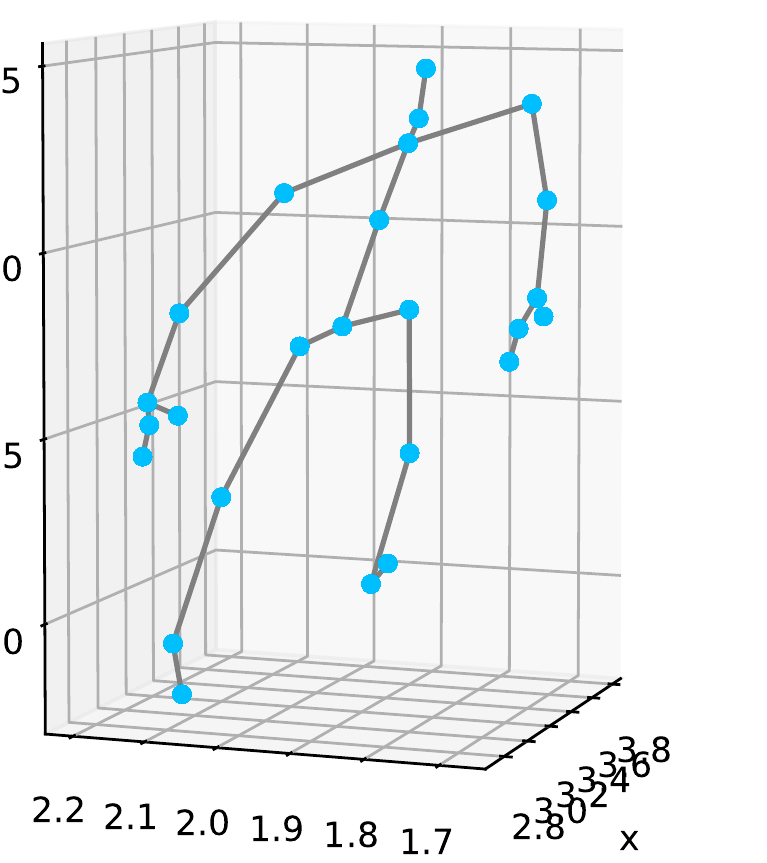}}
		\centerline{(c) shear-3}
	\end{minipage}
		%\end{tabular}
	\caption{Visualization of shear augmentation.}
	\label{shear}
\end{figure}

The shear transformation simulates the possible morphological changes of the skeleton caused by the system noise during the collection process. It is a linear mapping, and each joint can be moved in a fixed direction. The 3D coordinate shape of human joints can be tilted at any angle. The visualization pictures are shown in the Fig. \ref{shear} and shear transformation matrix is defined as:
\begin{equation}
\label{shear_matrix}
S
=
\left[\begin{array}{ccc}   1 & s_1 & s_2 \\ s_3 & 1 & s_4 \\ s_5 & s_6 & 1 \end{array}\right]
\end{equation}
where $s_1,s_2,s_3,s_4,s_5,s_6$ are shear factors sampled randomly from [-1,1]. All joint coordinates of the raw skeleton sequence are transformed by the shear matrix matrices.
\subsection{Gaussian noise}
\begin{figure}[htbp]
	\centering
	%\begin{tabular}{cc}
	\begin{minipage}{0.3\linewidth}
		\centerline{\includegraphics[width=2.5cm,height=3.5cm, trim=20 35 220 60,clip]{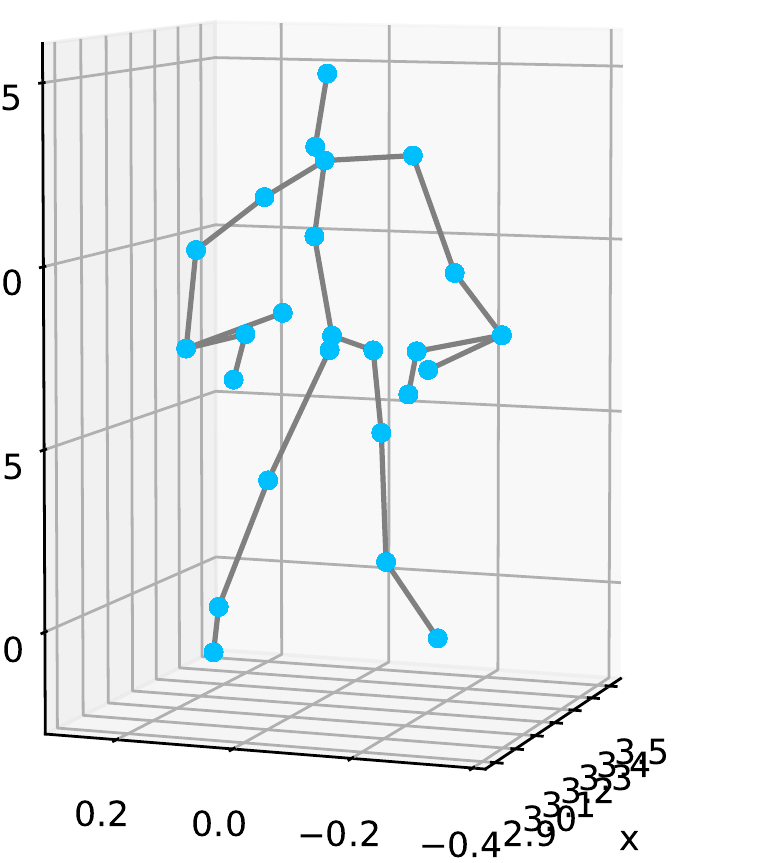}}
		\centerline{(a) Gaussion-1}
	\end{minipage}
	\hfill
	\begin{minipage}{.3\linewidth}
		\centerline{\includegraphics[width=2.5cm,height=3.5cm, trim=20 35 220 60,clip]{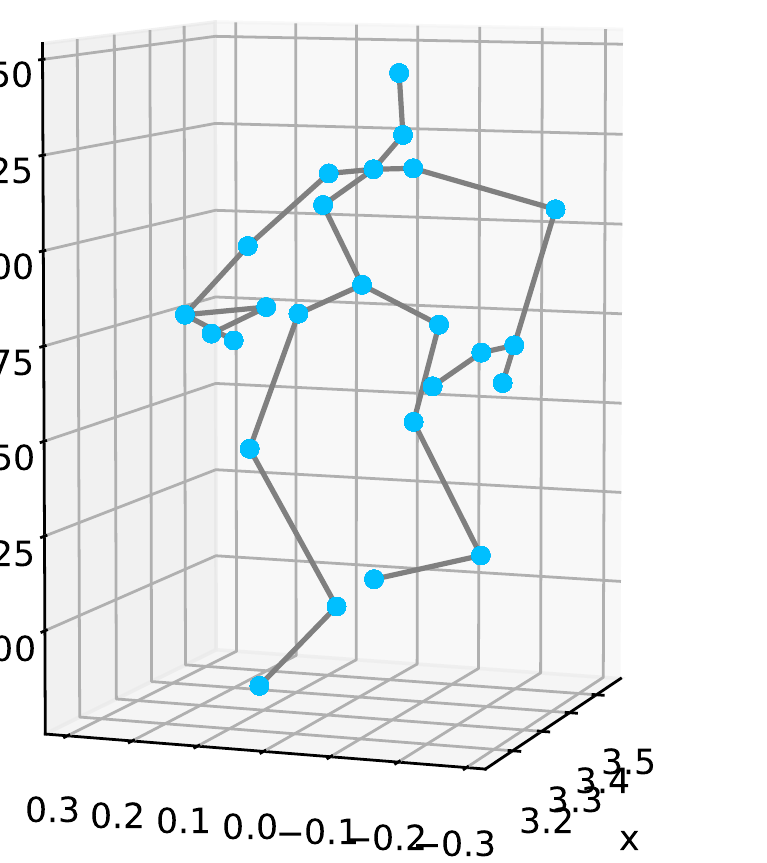}}
		\centerline{(b) Gaussion-2}
	\end{minipage}
	\hfill
	\begin{minipage}{.3\linewidth}
		\centerline{\includegraphics[width=2.5cm,height=3.5cm, trim=20 35 220 60,clip]{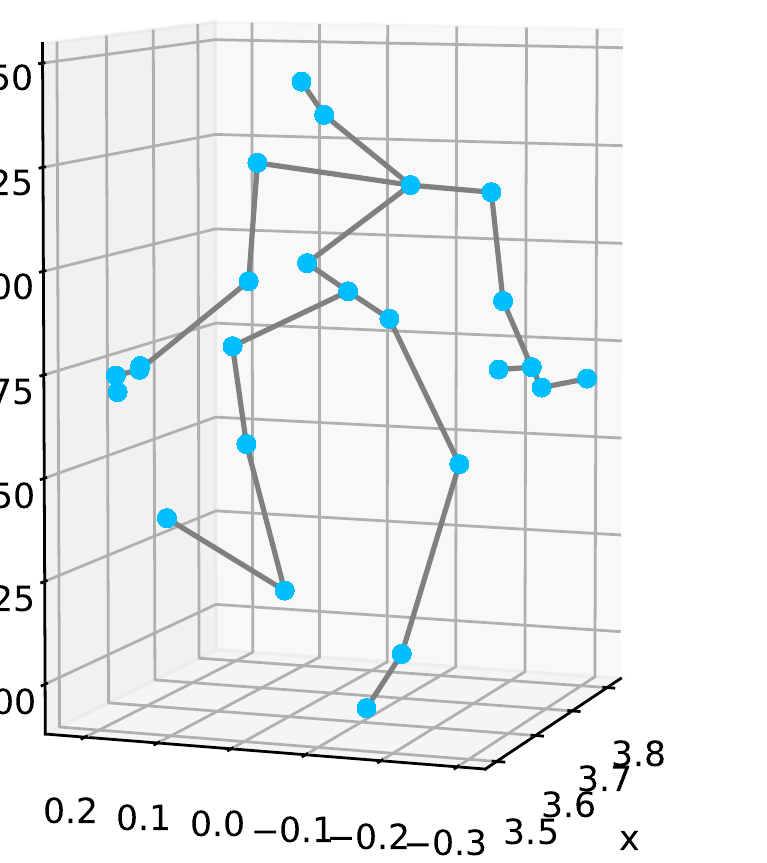}}
		\centerline{(c) Gaussion-3}
	\end{minipage}
		%\end{tabular}
	\caption{Visualization of Gaussion noise augmentation.}
	\label{Gaussion}
\end{figure}

Gaussian noise is an error in accordance with Gaussian normal distribution. In some cases, we need to add appropriate Gaussian noise to the standard data to make the data have certain errors. The visualization pictures are shown in Fig.  \ref{Gaussion}. 
\subsection{Joint mask}
\begin{figure}[htbp]
	\centering
	%\begin{tabular}{cc}
	\begin{minipage}{0.3\linewidth}
		\centerline{\includegraphics[width=2.5cm,height=3.5cm, trim=20 20 220 60,clip]{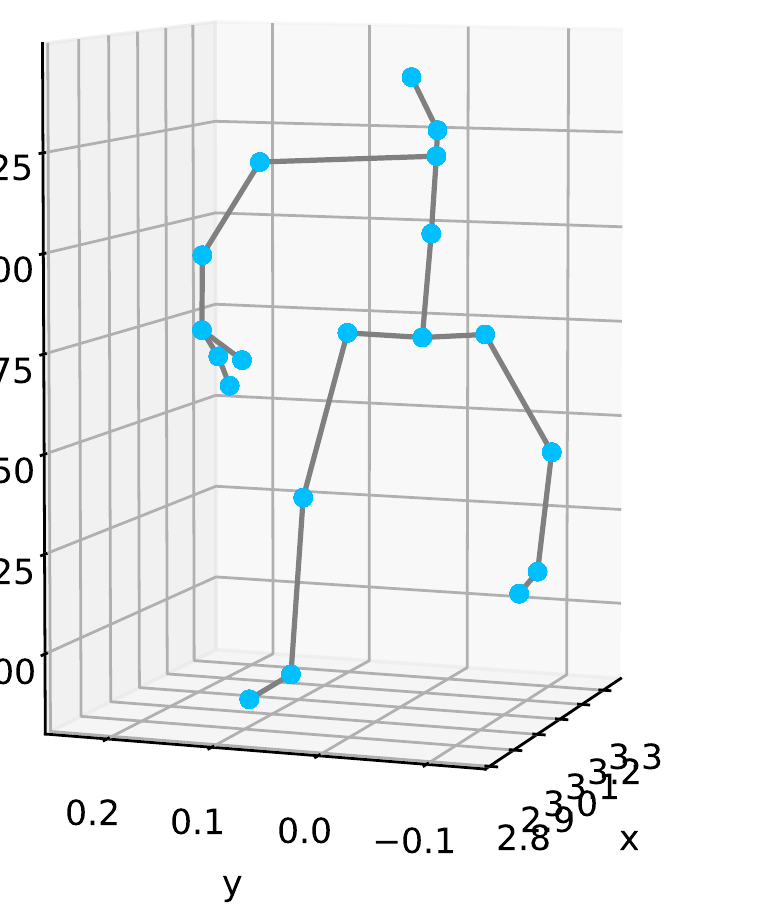}}
		\centerline{(a) joint-1}
	\end{minipage}
	\hfill
	\begin{minipage}{.3\linewidth}
		\centerline{\includegraphics[width=2.5cm,height=3.5cm, trim=20 20 220 60,clip]{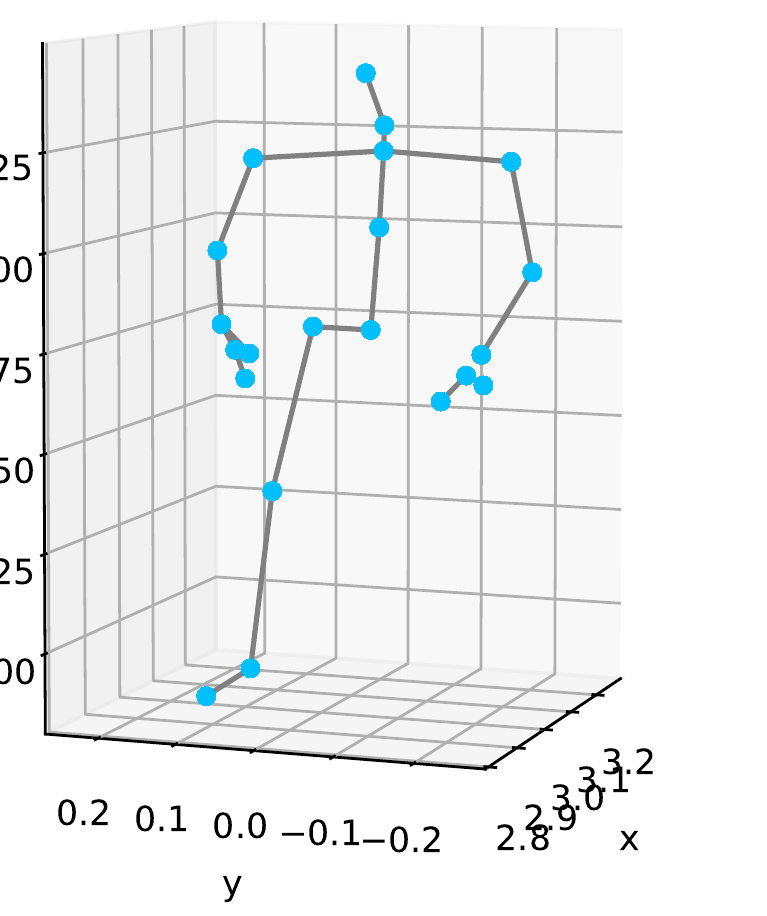}}
		\centerline{(b) joint-2}
	\end{minipage}
	\hfill
	\begin{minipage}{.3\linewidth}
		\centerline{\includegraphics[width=2.5cm,height=3.5cm, trim=20 20 220 60,clip]{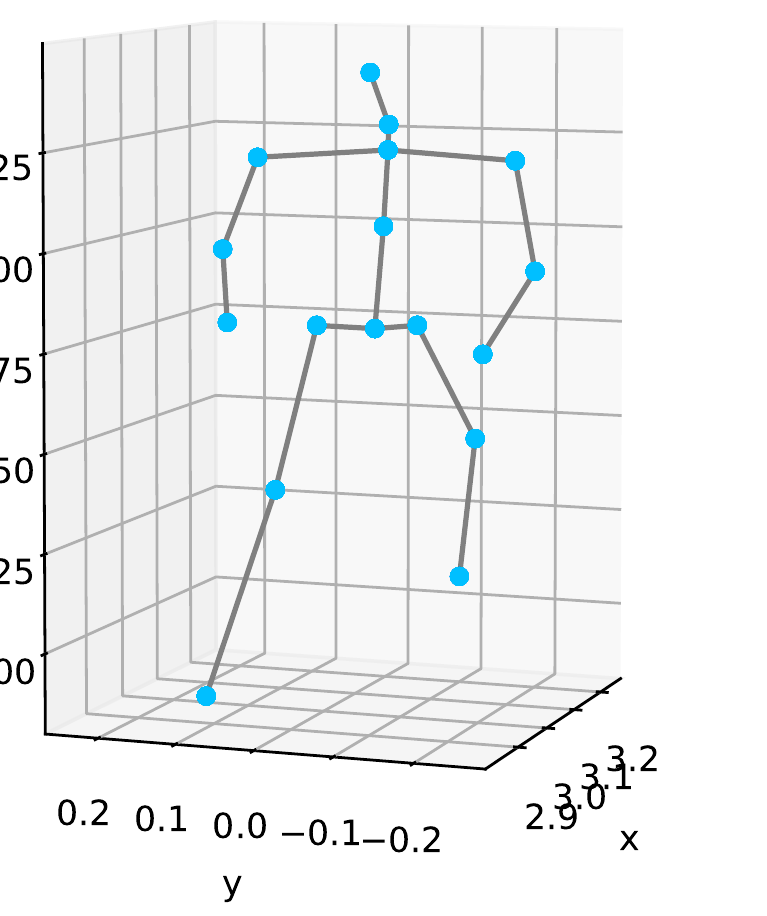}}
		\centerline{(c) joint-3}
	\end{minipage}
		%\end{tabular}
	\caption{Visualization of Joint Mask augmentation.}
	\label{Joint Mask}
\end{figure}

We employ the zero-mask (i.e., replace all coordinates by zeros) to a number of body joints in skeleton frames, which extends the pixel-level “Cutout” operation in image augmentation to joint-level skeleton sequences. This method simulates the situation where the body is obscured when collecting data, allowing the model to learn different local regions (except for the masked region) that probably contain crucial action patterns. More specifically, we randomly or specifically select body joints from random frames in the raw skeleton sequence to apply the zero-mask. The visualization pictures are shown in Fig. \ref{Joint Mask}. 

\subsection{Channel mask}
\begin{figure}[htbp]
	\centering
	%\begin{tabular}{cc}
	\begin{minipage}{0.3\linewidth}
		\centerline{\includegraphics[width=2.5cm,height=3.5cm, trim=20 20 220 60,clip]{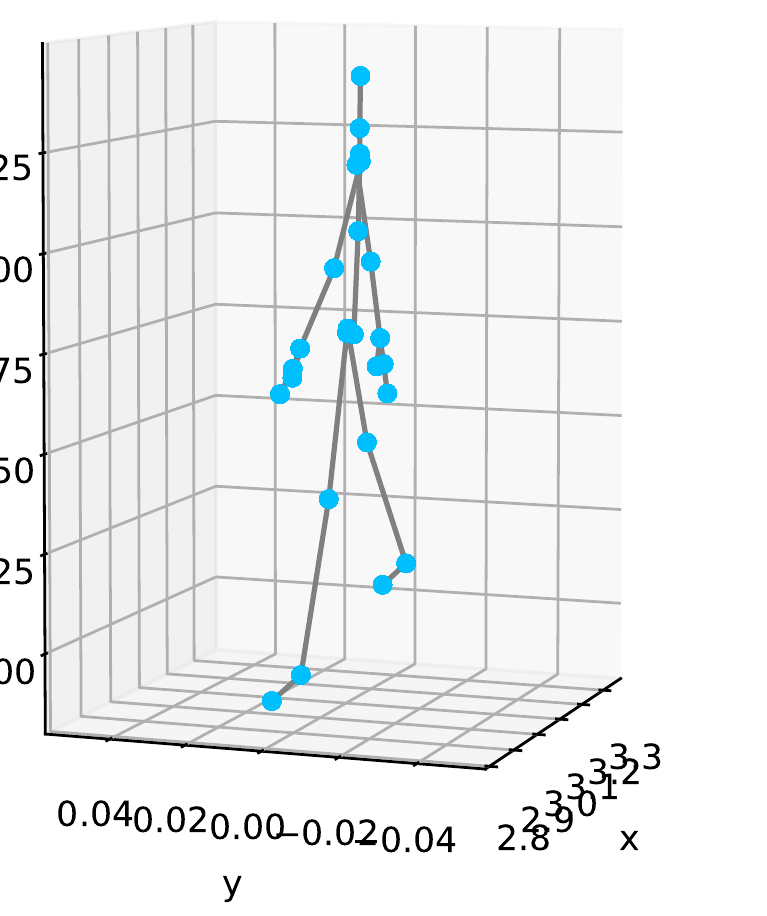}}
		\centerline{(a) channel-x}
	\end{minipage}
	\hfill
	\begin{minipage}{.3\linewidth}
		\centerline{\includegraphics[width=2.5cm,height=3.5cm, trim=20 20 220 60,clip]{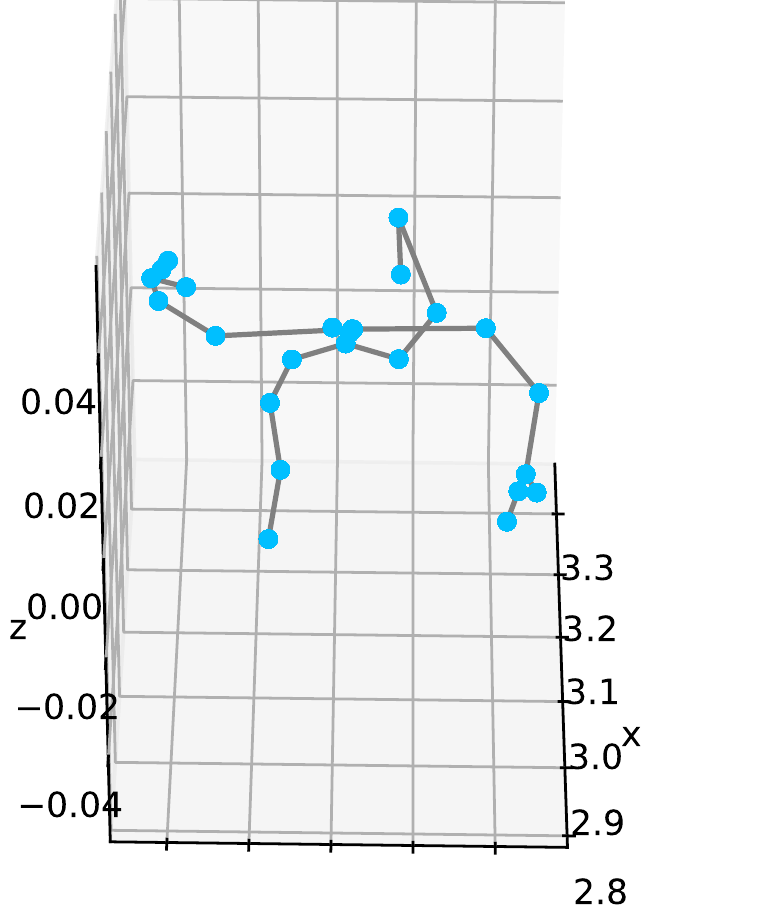}}
		\centerline{(b) channel-y}
	\end{minipage}
	\hfill
	\begin{minipage}{.3\linewidth}
		\centerline{\includegraphics[width=2.5cm,height=3.5cm, trim=20 20 220 60,clip]{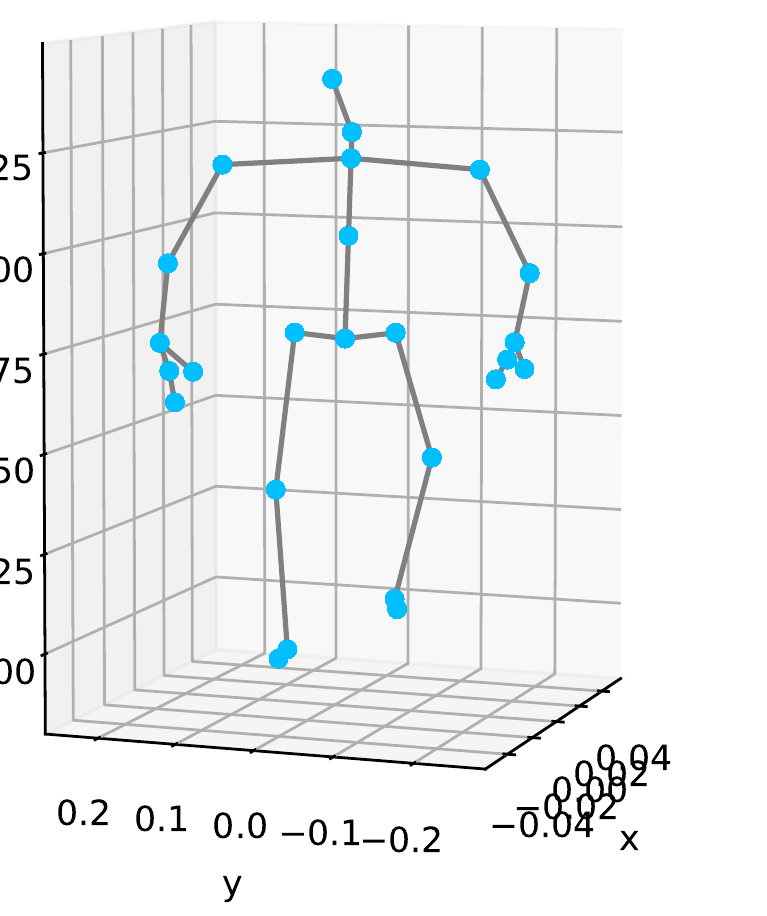}}
		\centerline{(c) channel-z}
	\end{minipage}
		%\end{tabular}
	\caption{Visualization of Channel Mask augmentation.}
	\label{channel Mask}
\end{figure}

The channel mask tries to simulate the skeleton at different viewing angles. We select a ''channel'' (an axis A $\in$ \{X, Y, Z\}) of the skeleton sequence separately, and apply the zero mask to all coordinates on this axis. In this way, the raw 3D skeleton sequence can be converted into a projected 2D sequence, so that the model can learn the changes in the human body's movement during walking from a specific two-dimensional plane. The visualization pictures are shown in Fig. \ref{channel Mask}. 

\section{Experiment Results and Discussion} \label{section:experimental}

\subsection{Classified method}
In this paper, we apply  two different deep learning neural networks to the augmented composite dataset to evaluate proposed data qugmentation methods by comparing their performance based on recognition rate. Because of the time series of gait skeleton data, we run two deep learning models (LSTM and TCN) under the framework of tensorflow$2.0$ in python $3.8$. The above classification methods are described below.
% we use two classification methods: LSTM \cite{hochreiter1997long}, and TCN \cite{lea2016temporal}. 

LSTM is composed of $7$-layer neural networks, including an input layer, two LSTM layers, two Dropout layers, a Flatten layer and a Dense layer. Two LSTM layers have $128$ and $512$ neurons respectively. In order to avoid overfitting, we set the dropout to $0.2$. The activation function is Tanh, and the loss function is the categorical crossentropy. The optimization algorithm of the model uses the Adam algorithm. A total of 200 epochs are trained, where the initial learning rate is $0.01$ and batch size is $64$.

TCN is composed of nine convolution layers (Convolution kernel = $8$), in which the stride of the fourth layer and the seventh layer is $2$, and the stride of the remaining layers is $1$. Each layer of the $1$st-$3$rd layers has $64$ neurons. Each layer of the $4$th-$6$th layers has $128$ neurons. Each layer of the $7$th-$8$th layers has $256$ neurons. The activation function is Relu and the dropout is $0.1$. The optimizer is Adam. The loss function is the categorical cross entropy, and the batch size is $32$.

We divided all participants' segments into a training set and a testing set according to the proportion of 4:1. In order to ensure the validity of the experimental results, the fragments are randomly divided by person, that is, fragments of the same person only appear in one group. On this basis, the training set and the test set are expanded to 5:1 and 3:1. Each augmentation method is tested on these three groups.

\subsection{Experimental results of rotation and translation augmentation} 
According to the rotation strategy proposed in the previous part of this paper, we carried out $11$ groups of experiments in the vertical and horizontal directions respectively. Table \ref{tab:rotation_h} and Table \ref{tab:rotation_v} show the classification results. One can see that from the data in Table \ref{tab:rotation_h} and Table \ref{tab:rotation_v}, the classification accuracy after rotation augmentation has been improved in some angles, the highest accuracy up to $92.15$\% (the data with $*$ in the table indicates that the accuracy after augmenting this angle is higher than the raw data). When the rotation angle is 18$\degree$ and 36$\degree$, the augmentation effect is significant, and the accuracy is improved in the three test sets. As the rotation angle increases, the difference between the raw data and the augmentation data is too large, so that the network cannot learn useful features. 

\begin{table*}
\setlength{\tabcolsep}{1.5mm}
\renewcommand\arraystretch{1.5} 
	\centering  
	\fontsize{6.5}{8}\selectfont 
\caption{Classification results of different rotation angles in the horizontal direction}
\label{tab:rotation_h}
\begin{tabular}{lllllllllllll}
\hline
                     &      & raw    & raw+18           & raw+36  & raw+54  & raw+72 & raw+90  & raw+108 & raw+126 & raw+144 & raw+162 & raw+180 \\ \hline
\multirow{2}{*}{3:1} & LSTM & 0.8431 & \textbf{0.8996*} & 0.8850* & 0.8442* & 0.8326 & 0.8576* & 0.8412  & 0.8631* & 0.8667* & 0.8759* & 0.84125 \\ \cline{2-13} 
                     & TCN  & 0.8686 & \textbf{0.9197*} & 0.9051* & 0.8099  & 0.8017 & 0.7973  & 0.8262  & 0.7973  & 0.8079  & 0.8332  & 0.8557  \\ \hline
\multirow{2}{*}{4:1} & LSTM & 0.7919 & \textbf{0.8101*} & 0.8081* & 0.7516  & 0.7452 & 0.7635  & 0.7896  & 0.8209* & 0.8205* & 0.8124* & 0.7694  \\ \cline{2-13} 
                     & TCN  & 0.7919 & \textbf{0.8751*} & 0.8334* & 0.7921  & 0.7914 & 0.8358* & 0.8138* & 0.7715  & 0.7961* & 0.7921  & 0.8123* \\ \hline
\multirow{2}{*}{5:1} & lSTM & 0.8430 & \textbf{0.8978*} & 0.8579* & 0.8065  & 0.8045 & 0.8073  & 0.8394  & 0.8502* & 0.8047  & 0.8175  & 0.8467* \\ \cline{2-13} 
                     & TCN  & 0.8540 & \textbf{0.8923*} & 0.8226  & 0.8089  & 0.8061 & 0.8010  & 0.8116  & 0.7700  & 0.7835  & 0.8357  & 0.8593* \\ \hline
\end{tabular}
\end{table*}

\begin{table*}[]
\setlength{\tabcolsep}{1.5mm}
\renewcommand\arraystretch{1.5} 
	\centering  
	\fontsize{6.5}{8}\selectfont 
\caption{Classification results of different rotation angles in the vertical direction}
\label{tab:rotation_v}
\begin{tabular}{lllllllllllll}
\hline
                     &      & raw    & raw+18           & raw+36           & raw+54  & raw+72  & raw+90  & raw+108 & raw+126 & raw+144 & raw+162 & raw+180 \\ \hline
\multirow{2}{*}{3:1} & LSTM & 0.8431 & \textbf{0.9051*} & 0.8988*          & 0.8649* & 0.8428  & 0.8558* & 0.8485* & 0.8540* & 0.8631* & 0.8503* & 0.8923* \\ \cline{2-13} 
                     & TCN  & 0.8686 & \textbf{0.9215*} & 0.8832*          & 0.8047  & 0.8034  & 0.8083  & 0.8007  & 0.7937  & 0.8057  & 0.7956  & 0.8833* \\ \hline
\multirow{2}{*}{4:1} & LSTM & 0.7919 & \textbf{0.8486*} & 0.8065*          & 0.8125* & 0.7866  & 0.7682  & 0.8285* & 0.8446* & 0.8056* & 0.8183* & 0.7682  \\ \cline{2-13} 
                     & TCN  & 0.7919 & 0.8445*          & \textbf{0.8478*} & 0.8147* & 0.8279* & 0.8467* & 0.8547* & 0.7810  & 0.7925* & 0.7859  & 0.8302* \\ \hline
\multirow{2}{*}{5:1} & lSTM & 0.8430 & \textbf{0.8759*} & 0.8448*          & 0.8357  & 0.8314  & 0.8503* & 0.8510* & 0.8248  & 0.8284  & 0.8266  & 0.8029  \\ \cline{2-13} 
                     & TCN  & 0.8540 & 0.8649*          & \textbf{0.9069*} & 0.8305  & 0.8291  & 0.8175  & 0.8098  & 0.8503  & 0.7953  & 0.8489  & 0.8043  \\ \hline
\end{tabular}
\end{table*}

\subsection{Experimental results of shear augmentation}
The shear matrix in Eq. \ref{shear_matrix} contains six shear factors, all of which are randomly generated from [-1,1]. In this paper, three groups of random shear augmentation experiments are carried out. In each group of experiments, each segment uses a different augmentation factor. From the experimental results in Table \ref{tab:shear}, it can be seen that the classification accuracy is unstable after shear augmentation. The highest accuracy rate is $91.97$\% and the lowest is $71.53$\%. In $4:1$ case, the average accuracy of the three groups under the LSTM model is $83.17$\%, but the accuracy of the third group is $14$\% higher than that of the second group.

\begin{table}[]
\setlength{\tabcolsep}{1.5mm}
\renewcommand\arraystretch{1.5} 
	\centering  
	\fontsize{6.5}{8}\selectfont 
\caption{Classification results of Shear augmentation}
\label{tab:shear}
\begin{tabular}{ccccccc}
\hline
                     &      & raw    & \multicolumn{1}{c}{1} & \multicolumn{1}{c}{2} & \multicolumn{1}{c}{3} \\ \hline
\multirow2*{3:1} & LSTM & 0.8430 & 0.8511                & 0.8105                & 0.9197                \\ \cline{2-6} 
                     & TCN  & 0.8686 & 0.8777                & 0.8323                & 0.7943                \\ \hline
\multirow2*{4:1} & LSTM & 0.7919 & 0.8029                & 0.7726                & 0.9197                \\ \cline{2-6} 
                     & TCN  & 0.7919 & 0.7992                & 0.8686                & 0.7153                \\ \hline
\multirow2*{5:1} & LSTM & 0.8430 & 0.8868                & 0.8014                & 0.8120                \\ \cline{2-6} 
                     & TCN  & 0.8540 & 0.8333                & 0.8467                & 0.7888                \\ \hline
\end{tabular}
\end{table}

\subsection{Experimental results of channel mask augmentation}

We perform four sets of experiments, and the results are shown in the Table \ref{tab:channel}. We find that after deleting the x-axis data, the classification accuracy is improved by more than $5$\%, reaching $91.34$\%; after deleting the z-axis data, it is greatly reduced.

\begin{table}[]
\setlength{\tabcolsep}{1.5mm}
\renewcommand\arraystretch{1.5} 
	\centering  
	\fontsize{6.5}{8}\selectfont 
\caption{Classification results of Channel Mask augmentation}
\label{tab:channel}
\begin{tabular}{llllll}
\hline
                     &      & raw    & raw+subx        & raw+xuby & raw+subz \\ \hline
\multirow2*{3:1} & LSTM & 0.8430 & \textbf{0.9035} & 0.8553   & 0.7689   \\ \cline{2-6} 
                     & TCN  & 0.8686 & \textbf{0.9134} & 0.8448   & 0.6587   \\ \hline
\multirow2*{4:1} & LSTM & 0.7919 & \textbf{0.8662} & 0.8196   & 0.6886   \\ \cline{2-6} 
                     & TCN  & 0.7919 & \textbf{0.8705} & 0.7973   & 0.6329   \\ \hline
\multirow2*{5:1} & LSTM & 0.8430 & \textbf{0.8981} & 0.8571   & 0.7379   \\ \cline{2-6} 
                     & TCN  & 0.8540 & \textbf{0.9095} & 0.8540   & 0.6313   \\ \hline
\end{tabular}
\end{table}

\subsection{Experimental results of adding Gaussian noise}
Noise may occur during data collection, transmission or processing, and noise is inevitable. In order to simulate the noise position that may appear in the above process, we add Gaussian noise N(0,0.05) to the joint coordinates of the raw sequence. The experiment results of three groups are shown in Table \ref{tab:gaussion}. After the first group of Gaussian augmentation, the accuracy of the two models increase to more than $90$\%, up to $93.43$\%, while the third group decreased significantly. The difference in classification accuracy between the two groups is more than $20$\%, indicating that the result of adding Gaussian noise is unstable.

\begin{table}[]
\setlength{\tabcolsep}{1.5mm}
\renewcommand\arraystretch{1.5} 
	\centering  
	\fontsize{6.5}{8}\selectfont 
\caption{Classification results of Gaussian augmentation}
\label{tab:gaussion}
\begin{tabular}{ccccccc}
\hline
                     &      & raw    & 1      & 2      & 3      \\ \hline
\multirow2*{3:1} & LSTM & 0.8430 & 0.9087 & 0.9288 & 0.7197 \\ \cline{2-6} 
                     & TCN  & 0.8686 & 0.9243 & 0.8099 & 0.7434 \\ \hline
\multirow2*{4:1} & LSTM & 0.7919 & 0.9343 & 0.9032 & 0.7281 \\ \cline{2-6} 
                     & TCN  & 0.7919 & 0.9343 & 0.7921 & 0.7921 \\ \hline
\multirow2*{5:1} & LSTM & 0.8430 & 0.9324 & 0.9069 & 0.7527 \\ \cline{2-6} 
                     & TCN  & 0.8540 & 0.9107 & 0.7989 & 0.7914 \\ \hline
\end{tabular}
\end{table}

\subsection{Experimental results of joint mask augmentation}

We explored the effects of removing the joints of the limbs, torso, upper body, and lower body on the recognition of depression. The classification results are shown in Table \ref{tab:joint_mask}. After removing any group of  data from the limbs, trunk, upper body and lower body, the accuracy decrease, indicating that the gait features of any part cannot be ignored. Then, add four groups of random trials and randomly delete the key nodes of $20$\%, $40$\%, $60$\%, $80$\%. After deleting too many joint points, the accuracy is greatly reduced.

\begin{table*}[]
\setlength{\tabcolsep}{1.5mm}
\renewcommand\arraystretch{1.5} 
	\centering  
	\fontsize{6.5}{8}\selectfont 
\caption{Classification results of Joint Mask augmentation}
\label{tab:joint_mask}
\begin{tabular}{cccccccccccc}
\hline
                     &      & raw    & raw+Upper\_body & raw+Lower\_body & raw+Trunk & raw+Limbs & 1      & 2      & 3       & 4      \\ \hline
\multirow2*{3:1} & LSTM & 0.8430 & 0.7923          & 0.8032          & 0.8288    & 0.7686    & 0.7916 & 0.9178 & 0.8087  & 0.7596 \\ \cline{2-11} 
                     & TCN  & 0.8686 & 0.7589          & 0.7975          & 0.7501    & 0.6812    & 0.7912 & 0.7828 & 0.85558 & 0.6624 \\ \hline
\multirow2*{4:1} & LSTM & 0.7919 & 0.7798          & 0.7080          & 0.7244    & 0.7116    & 0.8014 & 0.7812 & 0.6916  & 0.7135 \\ \cline{2-11} 
                     & TCN  & 0.7919 & 0.6770          & 0.7224          & 0.7088    & 0.6569    & 0.7708 & 0.8083 & 0.8188  & 0.8211 \\ \hline
\multirow2*{5:1} & LSTM & 0.8430 & 0.7651          & 0.7996          & 0.7406    & 0.8321    & 0.8080 & 0.9251 & 0.8130  & 0.7211 \\ \cline{2-11} 
                     & TCN  & 0.8540 & 0.7408          & 0.7927          & 0.6770    & 0.6441    & 0.7656 & 0.8010 & 0.8615  & 0.6694 \\ \hline
\end{tabular}
\end{table*}

\subsection{Discussion}
The above experiments prove that proposed augmentation strategies have a significant impact on the classification results. In order to analyze the reasons for the differences in classification accuracy of different strategies, we calculate the mutual information (MI) between the augmented data and the raw data. MI is an indicator to measure the "similarity" and it provides a general measure of dependence between variables \cite{steuer2002mutual}. The calculation formula of mutual information is Eq. \ref{MI}, which is derived from Eq. \ref{entropy}-\ref{union_entropy}. Eq. \ref{entropy} defines entropy, which is the average amount of information contained in each received message. The joint entropy H(A, B) of two discrete systems A and B is defined analogously in Eq. \ref{union_entropy}. 

\begin{equation}
\label{entropy}
\begin{array}{c}
H(A) = -\sum_{j = 1}^{M_{A}} p\left(a_{j}\right) \log p\left(a_{j}\right)
\end{array}
\end{equation}

\begin{equation}
\label{union_entropy}
\begin{array}{c}
H(A, B)=-\sum_{a,b} P_{A B}(a, b) \log P_{A B}(a, b)
\end{array}
\end{equation}

\begin{equation}
\label{MI}
\begin{array}{c}
I(A,B)=H(A)+H(B)-H(A,B)
\end{array}
\end{equation}

Table \ref{tab:MI} shows the average mutual information between each augmented dataset and the raw dataset. It can be seen from the table that the average  mutual information of data augmented by rotation and channal mask is larger, while it of the data augmented by the other three augmentation methods is relatively small. Combined with the principle of data augmentation, the data augmentation methods mentioned in this article can be divided into two categories. One is non-noise augmentation, that is, the basic shape of the skeleton is maintained during the data augmentation process, and the basic gait information is still retained. Such methods are mainly rotation and channel mask. The other is noise augmentation, that is, the skeleton structure changes in the process of data augmentation, leading to the deformation of the body. Such methods are mainly shearing and adding Gaussion noise.

\begin{table}[]
\setlength{\tabcolsep}{1.5mm}
\renewcommand\arraystretch{1.5} 
	\centering  
	\fontsize{6.5}{8}\selectfont 
\caption{Average Mutual Information of different augmentation methods in depression data}
\label{tab:MI}
\begin{tabular}{cccccc}
\hline
Rotation & Channel Mask & Gaussion Noise & Shear  & Joint Mask \\ \hline
6.4364   & 6.4207       & 5.1456         & 5.1438 & 4.9169     \\ \hline
\end{tabular}
\end{table}

From the data in Table \ref{tab:rotation_h}- Table \ref{tab:gaussion}, it can be observed that the effect of non-noise augmentation is significantly better than that of noise augmentation. For example, when rotating the smaller angles of 18$\degree$ and 36$\degree$, experimental results of both horizontal and vertical are significantly improved in the three groups of ratios. When the vertical rotation is 36$\degree$, the highest accuracy is $92$\%, and the accuracy of other angles is improved by at least $2$\%. As the angle increase, the difference between the raw data and the augmented data becomes larger and the classification accuracy shows a downward trend. In all 6 groups of results for deleting the x-axis, the accuracy of the channel mask is improved by at least $5$\%. These two augmentation methods are non-noise augmentation, and both maintain the basic shape of the skeleton and the raw gait information. However, because the augmentation process of shearing and adding Gaussian noise is random, the structure of the skeleton changes after the augmentation and the classification effect is unstable.

According to \cite{2017Differential}, adding appropriate data noise can improve the stability of the network and reduce the overfitting of training, but too much noise will mask the raw information of the data, thus reducing the accuracy of the model. Rotation augmentation and channel mask augmentation add appropriate noise to the raw data to improve the robustness of the model, while shear and Gaussian noise add too much noise, resulting in the network to learn a lot of biased data information in the training process.

% \begin{figure}[htbp]
% 	\centering
% 	%\begin{tabular}{cc}
% 	\begin{minipage}{0.48\linewidth}
% 		\centerline{\includegraphics[width=4.5cm,height=6.5cm, trim=20 20 220 30,clip]{fig_loss_raw.pdf}}
% 		\centerline{(a) raw data}
% 	\end{minipage}
% 	\hfill
% 	\begin{minipage}{.48\linewidth}
% 		\centerline{\includegraphics[width=4.5cm,height=6.5cm, trim=20 20 220 30,clip]{fig_loss_augmented.pdf}}
% 		\centerline{(b) shear augmentation}
% 	\end{minipage}
% 	\caption{Model curve display diagram.}
% 	\label{fig:loss}
% \end{figure}

\begin{figure}[htbp]
  \centering
  \centerline{\includegraphics[width=3.5in]{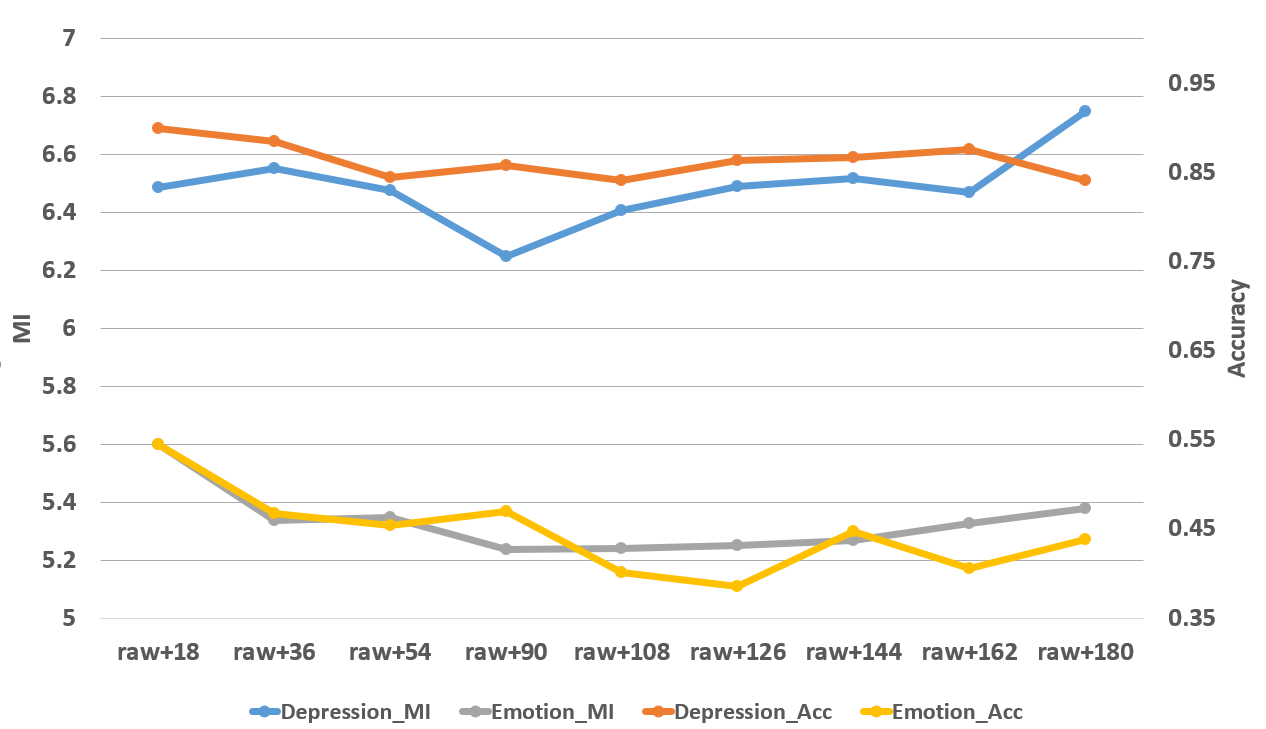}}
  \caption{Comparison chart of rotation augmentation and accuracy in the horizontal direction}
  \label{fig:Rotation_h_MI}
 \end{figure}

\begin{figure}[htbp]
  \centering
  \centerline{\includegraphics[width=3.5in]{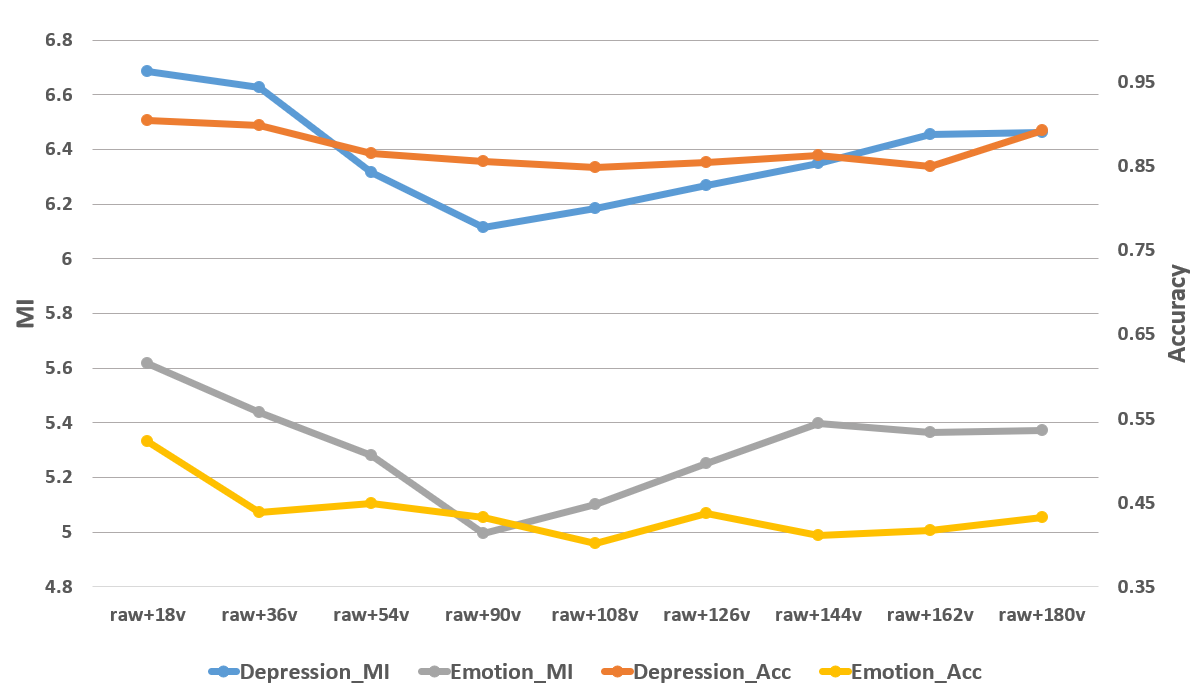}}
  \caption{Comparison chart of rotation augmentation and accuracy in the vertical direction}
  \label{fig:Rotation_v_MI}
 \end{figure}

In order to verify the above conclusions, we did the same experiment using the LSTM model on the emotion dataset mentioned above, and randomly selected 70 fragments of $2$ participants as the test set and the other $11$ participants as the training set. The mutual information and LSTM classification accuracies between each augmented dataset and raw dataset in the rotation augmentation are given in Fig. \ref{fig:Rotation_h_MI} and Fig. \ref{fig:Rotation_v_MI}. When rotating at a small angle in the vertical and horizontal directions, they all have higher mutual information and higher classification accuracy. The results of other augmentation methods are shown in the Table \ref{tab:emotion_MI}. After deleting the x-axis data in channel mask augmentation, the accuracy of the model is improved; After deleting the z-axis, the accuracy is still lower than the raw data, and the corresponding mutual information of the data after deleting the x-axis is greater than the data after deleting the z-axis. Random experiments are carried out for three augmentation methods: shear, Gaussian noise and joint mask. Similar to the depression data, the experimental results in each group are not stable.

\begin{table*}[]
\setlength{\tabcolsep}{1mm}
\renewcommand\arraystretch{1.5} 
	\centering  
	\fontsize{6.5}{8}\selectfont 
\caption{Accuracy and mutual information results of different augmentation methods}
\label{tab:emotion_MI}
\begin{tabular}{cccccccccccccccl}
\hline
\multicolumn{2}{c}{\multirow{2}{*}{}}                                             & \multirow{2}{*}{Raw} & \multicolumn{3}{c}{Channel Mask} & \multicolumn{3}{c}{Gaussion Noise} & \multicolumn{3}{c}{Shear} & \multicolumn{4}{c}{Joint Mask}    \\ \cline{4-16} 
\multicolumn{2}{c}{}                                                              &                      & raw+subx  & raw+suby  & raw+subz & 1          & 2         & 3         & 1       & 2      & 3      & 1      & 2      & 3      & \multicolumn{1}{c}{4}      \\ \hline
\multirow{2}{*}{\begin{tabular}[c]{@{}c@{}}Depression \\ Data\end{tabular}} & MI  & -                    & 6.5910    & 6.5113    & 6.1598   & 5.3136     & 5.0620    & 5.0612    & 5.2258  & 5.1094 & 5.0960 & 6.1348 & 6.1075 & 5.1751 & 4.2244 \\ \cline{2-16} 
                                                                            & ACC & 0.7919               & 0.8662    & 0.8196    & 0.6886   & 0.9343     & 0.9032    & 0.7281    & 0.8029  & 0.7726 & 0.9197 & 0.7814 & 0.7812 & 0.6916 & 0.7135 \\ \hline
\multirow{2}{*}{\begin{tabular}[c]{@{}c@{}}Emotion \\ Data\end{tabular}}    & MI  & -                      & 4.218     & 4.0167    & 3.2506   & 3.9731     & 2.9616    & 3.9738    & 3.1721  & 3.5317 & 3.1372 & 3.3829 & 4.1910 & 2.7314 & 2.9880 \\ \cline{2-16} 
                                                                            & ACC & 0.4188               & 0.6667    & 0.5833    & 0.3846   & 0.6875     & 0.3646    & 0.5677    & 0.4948  & 0.5365 & 0.3750 & 0.4844 & 0.7945 & 0.3802 & 0.3437 \\ \hline
\end{tabular}
\end{table*}

These values enable us to explore the relationship between the mutual information of gait skeleton information generated from an augmentation strategy and the test accuracy of LSTM trained on that augmented training set. In other words, we attempt to determine if the extent to which an augmentation captures information about the raw training set affects the classification accuracies. Looking at the results in the figure, we notice that when using the non-noise augmentation methods, higher mutual information is correlated with higher classification accuracy. Specifically, the augmentations that result in data have mutual information of approximately $6.3-6.4$ (rotation, channel mask) have accuracies of around $0.8$ and the accuracy varies with the rise and fall of the mutual information. While the accuracy of the augmentation data (shear, Gaussian noise) with an average mutual information of about $5.1$ is quite unstable. 

Combining the above analysis, our conclusion is the augmented training data set that retains more of the raw skeleton data properties determines the performance of the detection model. Small angle rotation and x-axis have a large amount of mutual information between the augmented data and the raw data, indicating that it greatly retains the raw skeleton information, and the corresponding classification accuracy is higher. However, shear, Gaussian noise and joint mask augmentation have little mutual information between the augmented data and the raw data. These augmentation methods deform the raw skeleton, which adds more noise to the training data and seriously affects the accuracy of classification.

Generally, this paper proposes five depression data augmentation methods and explores the effectiveness of these five methods in combination with mutual information. Compared with previous studies, this is the first time that the data augmentation method has been applied to the gait dataset. It solves the problem that the scene is difficult to cover in the process of gait data collection, the collection angle is single, and the neural network is easy to overfit. However, the following problems also existed in the research process. First of all, only the five augmentation methods are discussed separately, and there is no further discussion on the combination method. Second, accuracy. At present, the best result of this paper can reach $92.15$\%, which is far from the actual application and needs to be further improved.
 
\section{Conclusions and future work} \label{conclusion}
In this paper, we propose five methods to augment depression skeleton data, and further explore the effectiveness of those. By comparing the classification results of the two datasets and the mutual information between the augmented data and the raw data, the augmentation methods are divided into non-noise augmentation and noise augmentation, and the effect of non-noise augmentation is obviously better than noise augmentation. Then, combined with the principles of various augmentation methods, we conclude that the augmented training data set that retains more of the raw skeleton data properties determines the performance of the detection model.

In future work, we will continue to explore other augmentation strategies and combine neural networks to achieve faster and more accurate depression recognition. At the same time, we will further study the relationship between gait and depression, and explore more efficient and more suitable depression screening methods.

%\begin{acknowledgements}
%\end{acknowledgements}


\bibliographystyle{unsrt}
\bibliography{refs}   % name your BibTeX data base

\end{document}